\pdfoutput=1

\documentclass[11pt]{article}

\usepackage{acl}

\usepackage{times}
\usepackage{latexsym}

\usepackage[T1]{fontenc}

\usepackage[utf8]{inputenc}

\usepackage{microtype}
\usepackage{bm} 
\usepackage{amsmath}
\usepackage{amssymb}
\usepackage{multirow}
\usepackage{multicol}
\usepackage{booktabs}
\usepackage{graphicx}
\usepackage{xcolor}
\usepackage{bbding}
\usepackage{siunitx}
\usepackage{tabularx}
\usepackage{hyperref}
\usepackage{authblk}
\usepackage{inconsolata}

\title{{\sc MPCoder}: Multi-user Personalized Code Generator with Explicit and Implicit Style Representation Learning}

\author{\bf ~~Zhenlong Dai$^{1}$, 
        ~~Chang Yao$^{1}$, 
        ~~Wenkang Han$^{1}$, 
        ~~Ying Yuan$^{2}$, 
        ~~Zhipeng Gao$^{1}$\thanks{Co-corresponding authors.},  
        ~~Jingyuan Chen$^{1*}$\\
        \textsuperscript{1}Zhejiang University \\
        \textsuperscript{2}Zhejiang Police College \\
        {\tt\small \{zhenlongdai, changy, wenkanghan, tracy1108, zhipeng.gao, jingyuanchen\}@zju.edu.cn} 
        }

\begin{document}
\maketitle

\begin{abstract}

Large Language Models (LLMs) have demonstrated great potential for assisting developers in their daily development. 
However, most research focuses on generating correct code, how to use LLMs to generate personalized code has seldom been investigated. 
To bridge this gap, we proposed {\sc \textbf{MPCoder}} (\textbf{\underline{M}}ulti-user \textbf{\underline{P}}ersonalized \textbf{\underline{Code}} Generato\textbf{\underline{r}})  to generate personalized code for multiple users. 
To better learn coding style features, we utilize explicit coding style residual learning to capture the syntax code style standards and implicit style learning to capture the semantic code style conventions. 
We train a multi-user style adapter to better differentiate the implicit feature representations of different users through contrastive learning, ultimately enabling personalized code generation for multiple users. 
We further propose a novel evaluation metric for estimating similarities between codes of different coding styles. 
The experimental results show the effectiveness of our approach for this novel task. 
 
The code and dataset are available at \url{https://github.com/455849940/MPCoder}.

\end{abstract}

\section{Introduction}

\begin{figure}
\includegraphics[width=0.48\textwidth]{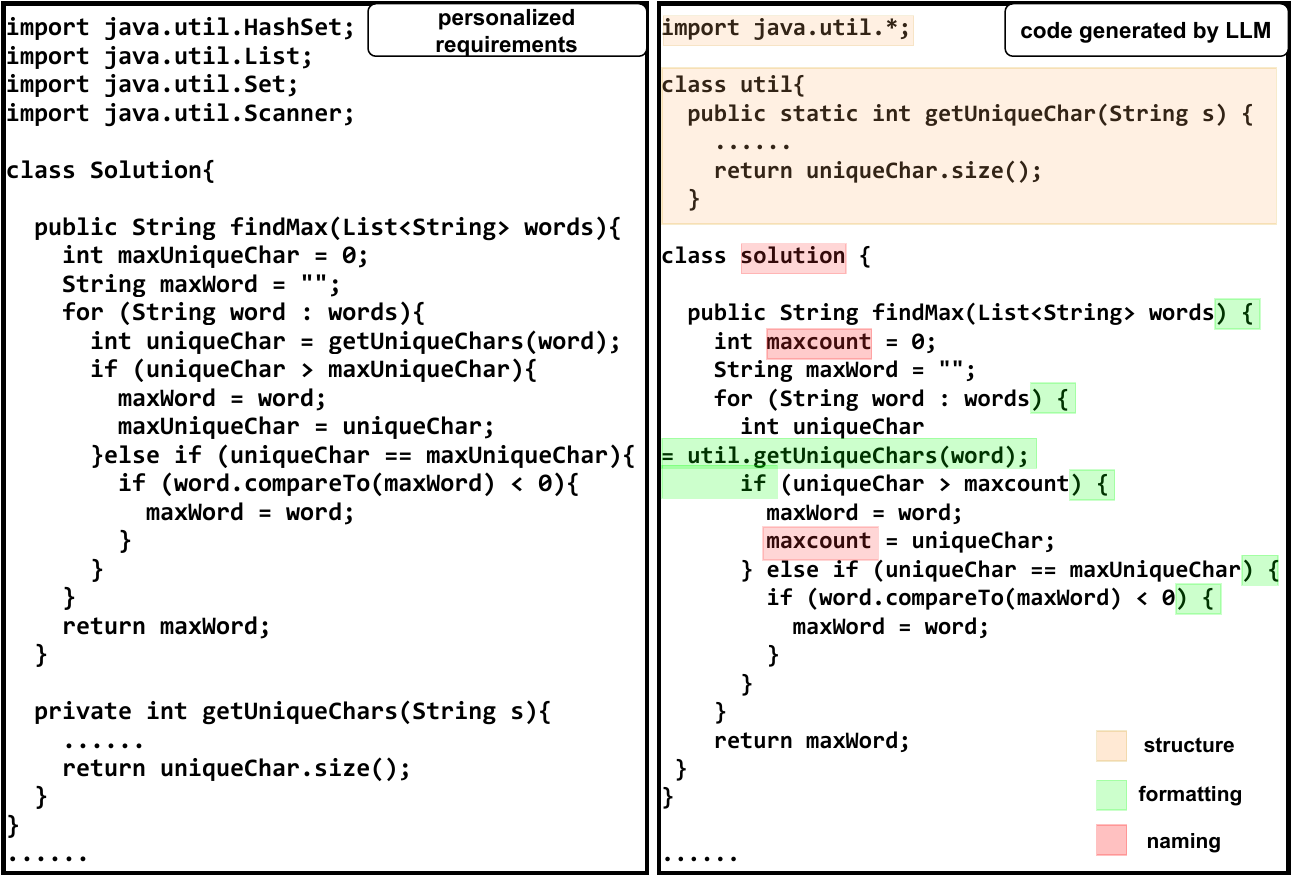}
\vspace{-2em}
\caption{Example of code generated by LLMs and the corresponding personalized code that is expected, with areas inconsistent with the expectations marked in different colors within the model-generated code. 
\vspace{-2em}
}
\label{fig:Compare}
\end{figure}

Nowadays, LLMs have been successfully used to support developers' daily development, such as code generation, test generation, etc. 
However, existing Code LLMs are usually general models trained with large programming corpus~\cite{zheng2023codegeex, chen2022cat}, therefore the generated code is difficult to adapt to personalized and/or customized requests. 
Consider the following practical scenarios:
Alice is a software developer. To improve programmers' daily efficiency, her company provided the base LLMs that can be used for code generation. 
Nonetheless, different developers/projects have their own coding standards and specifications. 
If Alice needs to generate code satisfying specific conventions, the base LLMs may fail to capture these nuanced differences. 
As shown in Fig.~\ref{fig:Compare}, Alice has to painstakingly revise and review the generated code. If the custom style of the generated code conforms to the standard style of different developers/projects and is correct, it can greatly increase developer productivity~\cite{kropp2013teaching, cheng2022improves,song2023artificial} and reduce code maintenance costs~\cite{alkhatib1992maintenance, tu2014localness}.

Recent researchers have explored code generation task by using LLMs; however, most studies~\cite{li2023starcoder,li2022competition, ahmad2021unified,hu2021lora} focus on generating ``correct'' code. There is limited research investigating how to generate ``personalized'' code, especially for multi-user personalization, with no research conducted yet. 
Automatically generating code according to developers' preferences or projects' consistency is a challenging task:  
(i) Considering different programmers have their own coding styles, \textbf{it is too expensive to fine-tune an LLM for each user}~\cite{guo2021text}. 
Therefore, how to build an efficient model to generate personalized code for multiple users poses a significant challenge. 
(ii) \textbf{Coding styles are hard to learn and capture.}
Coding styles include different aspects of the code, such as code naming, formatting, and structures. 
How to distinguish coding style differences between different users and obtain good style representations is another challenge. 
(iii) \textbf{Coding styles are hard to evaluate.}
Unlike code correctness, which can be evaluated by executing test cases, there is no evaluation metric to estimate coding styles of different code fragments. 
How to evaluate coding styles quantitatively becomes another challenge for our study~\cite{husain2019codesearchnet,lu2021codexglue}.

To tackle the above challenges, we propose a novel approach named {\sc \textbf{MPCoder}}, which is designed to generate personalized code for multiple users according to their individual coding styles. 
After training, our model can be easily queried with the ID of the user and generate personalized code consistent with his/her desired coding styles.  
To better capture styles within the raw code, we encode the \textbf{explicit style features} and \textbf{implicit style features} to obtain an effective coding style representation. 
Regarding the explicit style features, we apply the coding style checking tool (Checkstyle\footnote{\label{checkstyle}http://checkstyle.sourceforge.net} in our study) to detect different coding style attributes explicitly, then the model learn and encode these style attributes by residual learning, which guides the model in identifying the coding style attribute by contrasting two sets of coding style attributes (Section~\ref{sec:ExplicitLearning}). 
For the implicit style features, to capture the subtle and unnoticed style differences between different users, we design a multi-user style adapter to further distinguish coding style differences among different users by using contrastive learning (Section~\ref{sec:ImplicitLearning}). 
 
Finally, by combining the explicit coding style features and user-specific implicit style features, we can generate the user's personalized code that both contain the syntax and semantic styles 
of the code (Section~\ref{sec:Two-train}). 
Due to the limited prior work on exploring personalized code generation, there is currently no effective way to estimate whether two pieces of code have similar coding styles.
In this paper, we propose a novel evaluation metric, \textbf{Coding Style Score} (CSS), to quantitatively estimate the coding styles between two given codes.

In summary, our paper makes the following contributions: 
Firstly, current research mainly focuses on generating correct code, to the best of our knowledge, no prior work explored how to generate personalized code for multiple users.  
Secondly, we develop a novel model, named {\sc MPCoder}, to learn multi-users' coding style features and generate personalized code. 
Thirdly, we propose a novel evaluation metric for estimating coding styles quantitatively, and additionally, we also release our dataset which contains source code written by multiple users.  
The experimental results show the effectiveness of our model over a set of baselines, showing its ability to generate personalized code while minimizing the degradation of code correctness. 
We hope our study can lay the foundations for this research topic and provide valuable insights into the potential for personalized generation capabilities of general LLMs.

\begin{figure*}[t]
\centering
\includegraphics[width=0.95\textwidth]{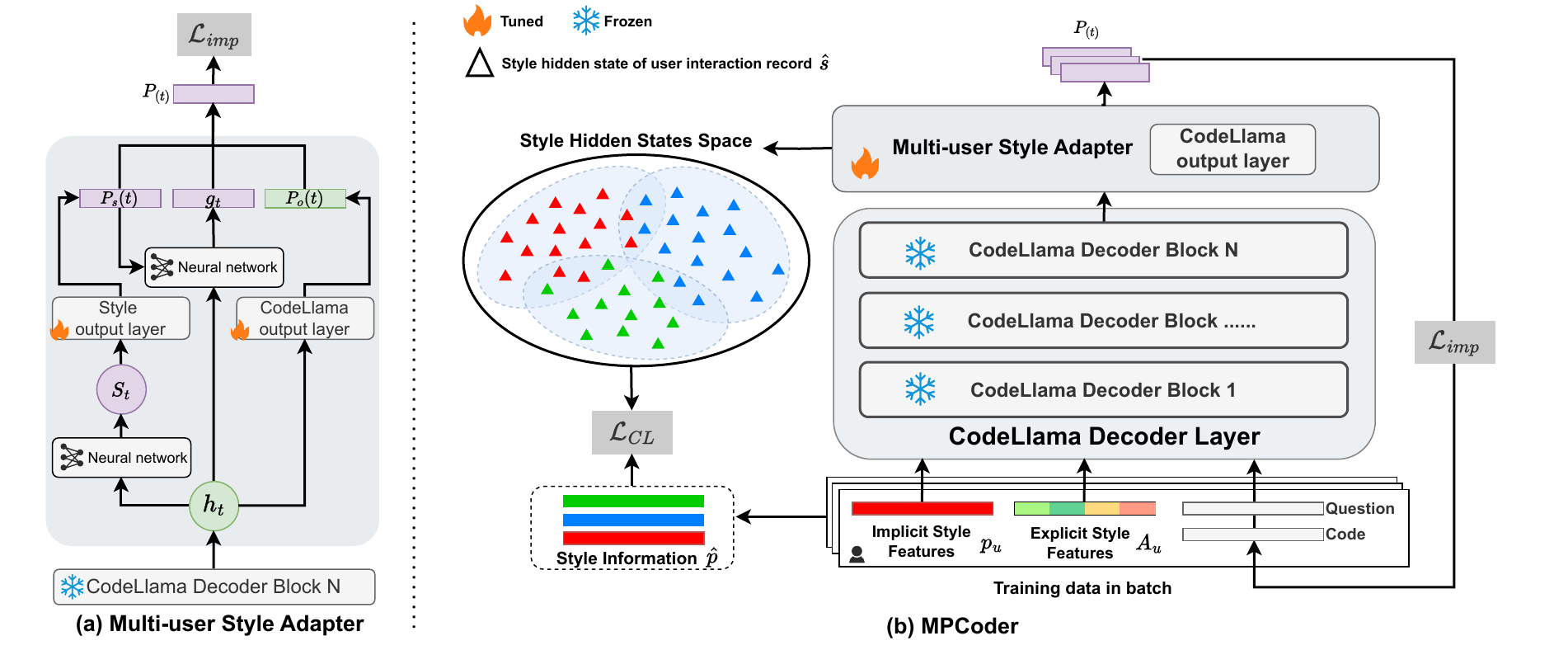}
\vspace{-1em}
\caption{Overview of {\sc MPCoder}. (a) illustrates the structure of the multi-user style adapter. (b) is the second training stage of {\sc MPCoder} at the decoding step \(t\).
} 

\label{fig:model}
\vspace{-1em}
\end{figure*}

\vspace{-0.5em}
\section{Methodology}
The coding style can be categorized into \textit{syntax style} and \textit{semantic style}, according to the structure and meaning of the code. 
Syntax style refers to the formatting rules of the code (\textit{e.g.},  indentation, spacing, capitalization of variables) which can be easily defined~\cite{allamanis2014learning,markovtsev2019style}. 
On the other hand, semantic style refers to the use of language features and constructs to convey intent (\textit{e.g.}, design patterns, and meaningful names of code) which is hard to precisely define in language~\cite{parr2016towards,ogura2018bring}.
Both syntax and semantic style are important for making the code more readable and maintainable. Details of the syntax and semantic coding style differences can be found in Appendix~\ref{subsec:B3}.

To capture these two types of coding styles, as illustrated in Fig.~\ref{fig:model}, we propose a novel approach {\sc MPCoder}, which utilizes explicit coding style learning (Section~\ref{sec:ExplicitLearning}) to capture the syntax style standards pre-defined by industry and implicit coding style learning (Section~\ref{sec:ImplicitLearning}) to capture the semantic style that is learned from the code itself. Moreover, a multi-user style adapter (Section~\ref{sec:ImplicitLearning}) is trained to estimate the style probability distribution for multiple users. After two stages of training , {\sc MPCoder} can generate personalized code for multiple users simultaneously(Section~\ref{sec:Two-train}).

\subsection{Task Definition}
\label{sec:task-def}
Given a specific programming question $q$, the task of multi-user personalized code generation is to generate the corresponding code $c$ for a particular user $u\in U$ based on his/her historical programming records $r$, where the generated code $c$ should be consistent with the user's historical coding styles.  It is important to note that the model should be able to generate personalized code for different users simultaneously.

More formally, the objective of personalized code generation is to learn the underlying conditional probability distribution $P_{\theta}(c|r,q,u)$ parameterized by $\theta$. 
In other words, the goal is to train a model $\theta$ such that the probability $P_{\theta}(c|r,q,u)$ is maximized over the training dataset in order to generate personalized responses. 
\begin{figure}[t]
\centering
\includegraphics[width=0.48\textwidth]{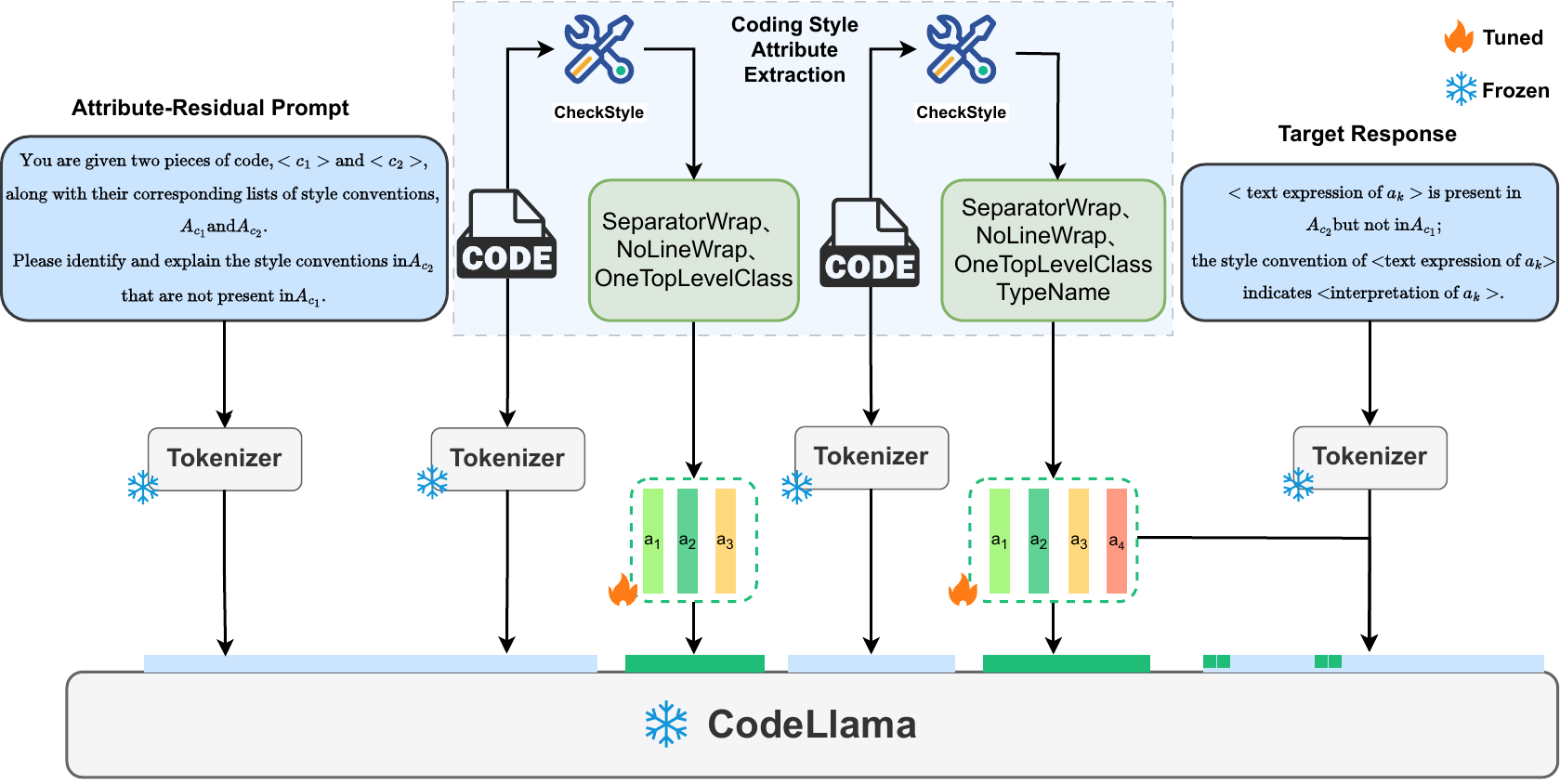}
\vspace{-1em}
\caption{Explicit coding style residual learning.}
\label{fig:Redisual_Learning}
\vspace{-2em}
\end{figure}

\vspace{-1em}
\subsection{Explicit Coding Style Learning}
\label{sec:ExplicitLearning}
\textbf{Coding Style Attribute Extraction.} The syntax style of code refers to a set of guidelines on how code is organized, such as indentation, spacing, capitalization of variables. 
These coding styles can be explicitly identified using a standard coding style checking tool. In this study, we use the Checkstyle\footref{checkstyle} tool to explicitly examine 25 coding style attributes (\textit{e.g.}, \texttt{SeparatorWrap}, \texttt{NoLineWrap}, etc.), with the detailed information provided in Appendix~\ref{subsec:A3}.

Specifically, a piece of code $c$ is identified by the Checkstyle tool to contain multiple coding style attributes. These attributes are vectorized as learnable representations, denoted as $A_c=\{a_1,a_2,...,a_k\}$. Here, $k$ is the number of attributes, $a_i\in \mathbb{R}^H$ denotes the $i$-th attribute where each $a_i$ belongs to the 25 coding style attributes defined by Checkstyle, and $H$ represents the dimension of the representation which is the same as the word embedding dimension of LLMs.

\noindent\textbf{Coding Style Residual Learning.} 

It is challenging for LLMs to directly acquire and distinguish the representations of all coding style attributes from the corresponding code. Inspired by previous study~\cite{alayrac2022flamingo}, which suggests that providing LLMs with two similar images along with their differences simultaneously can enhance their learning of image representations, we propose a novel residual learning mechanism to aid in the explicit recognition of each coding style attribute. Given two code fragments with similar coding style attributes, we guide the LLM in identifying the residual attribute to learn more precise and distinguishable attribute representations, as shown in Fig.~\ref{fig:Redisual_Learning}.

Specifically, two pieces of code $c_{1}$ and $c_{2}$, exhibit similar coding style attributes denoted by $A_{c_{1}} = \{a_1,a_2,...,a_{k-1}\}$ and $A_{c_{2}} = \{a_1,a_2,...,a_k\}$, where $a_k$ represents the residual attribute. Here we design attribute-residual prompt $R$, which inspires LLMs to identify the residual attribute $a_k$ by comparing the representations of $A_{c_{1}}$ and $A_{c_{2}}$ as:
\begin{figure}[!thb]
\centering
\resizebox{1.0\linewidth}{!}{%
\small
\fcolorbox{black}{gray!6}{%
\parbox{\linewidth}{%
\noindent$\bullet$ \textbf{Attribute-Residual Prompt.}
You are given two pieces of code, <$c_{1}$> and <$c_{2}$>, along with their corresponding lists of style conventions, $A_{c_{1}}$ and $A_{c_{2}}$. Please identify and explain the style conventions in $A_{c_{2}}$ that are not present in $A_{c_{1}}$.\\
\noindent$\bullet$ \textbf{Target Response.}
<text expression of $a_k$> is present in  $A_{c_{2}}$ but not in $A_{c_{1}}$; the style convention of <text expression of $a_k$> indicates <interpretation of $a_k$>.
}
}
}
\label{fig:residual-prompt}
\end{figure}

\noindent Here, <text expression of $a_k$> represents the name of attribute $a_k$, and <interpretation of $a_k$> denotes the detailed explanation.
Through residual training, the learned coding style attribute representations are consistent with the corresponding syntax styles.
\noindent\textbf{Training Objective.} 
In this training stage, the objective is to minimize the negative log-likelihood by utilizing the attribute-residual prompt as:
\begin{align}
\mathcal{L}_{\text{exp}}= -\sum_{t=1}^{l} \log P(x_t|x_{<t};A),
\end{align}
where \(l\) is the number of tokens in the attribute-residual prompt and the target response, $P \in \mathbb{R}^{|\mathcal{V}|}$ is the probability distribution on LLM's vocabulary, and $A$ denotes the representations of coding style attributes that are to be learned in this stage.

\subsection{Implicit Coding Style Learning}
\label{sec:ImplicitLearning}

Explicit coding style learning mainly focuses on capturing the syntax style features that are pre-defined by industry standards and are considered to be user-independent. 
However, it is important to note that there is another type of coding style features that are more difficult to precisely define using language. These features are user-specific and may vary from one individual to another. We define this feature as semantic style feature, which refers to the use of language features and constructs to convey intent.
For example, some user prefer to use $i,k,j$ to denote variables, while others may want to use more meaningful variable names.

To obtain semantic features of coding styles, we utilize implicit coding style learning to learn each user's style features from their historical coding records.
Since detecting the coding styles of various users can be intricate and challenging, we propose a multi-user style adapter to better differentiate the implicit feature representations of different users through contrastive learning. Subsequently, the multi-user style adapter estimates the probability distribution of styles over the entire vocabulary for multiple users. Additionally, it incorporates personalized fine-tuning based on implicit features, thereby minimizing the necessity to fine-tune and store multiple copies of LLM for different users.

\paragraph{Implicit Style Features.}
Inspired by personalized lightweight fine-tuning~\cite{lester2021power,li2023personalized,zlotchevski2022exploring}, we guide the personalized generation of LLMs for each user $u \in U$ with a set of pre-trained continuous vector representations, namely implicit style features. 
Specifically, we aim to learn implicit style features $p_u = \{p_1,p_2,...,p_m\}$ for user $u$ using their historical coding records. 
Here, $p_i\in \mathbb{R}^H$ is the $i$-th learnable representation for user $u$ and $m$ is the number of learnable representations. We keep most of the LLMs' parameters fixed, and only tune the output layer to accelerate convergence. Given a programming question $q$ and its corresponding code $c$, represented as a token sequence $x = \{x_1,x_2...,x_n\}$ of length $n$, we obtain a sequence of token embeddings $e$ through the embedding layer of LLMs. User-specific semantic style features $p_u$ are then concatenated with token embeddings as the input of decoder layer in LLMs as:
\begin{align}
e & = \texttt{EmbeddingLayer}(\{x_0,x_1...,x_n\}) ,\\
h & = \texttt{DecoderLayer}([p_u;e]),
\label{eq:x}
\end{align}
where $e = \{e_1,e_2,...,e_n\}$ and the hidden states of the decoder layer are denoted as $h=\{h_1,h_2,...,h_{n+m}\}$. Each token embedding $e_i$ and implicit style feature $p_i$ share the same dimension $H$.
Implicit semantic style features $p_u$ are learned based on the following prompt template:
\begin{figure}[!thb]
\centering
\resizebox{1.0\linewidth}{!}{%
\small
\fcolorbox{black}{gray!6}{%
\parbox{\linewidth}{%
\noindent$\bullet$ \textbf{Prompt template.}
<$p_u$>
Give you a programming question <$q$> and corresponding user coding style conventions <$A_u$>, please give the corresponding style of the answer in Java.\\
\noindent$\bullet$ \textbf{Target Code.}
<c>
}%
}%
}
\label{fig:input}
\end{figure}

\noindent where $A_u$ denotes the learned coding style attribute representations of user $u$ in the first stage.  
After the training process, these user-specific semantic style features $p_u$ can be learned from the user-generated code itself, allowing for the implicit expression of the users' coding style.

\paragraph{Multi-user Style Adapter.}

The generic nature of the output layer in LLMs entails that it does not take into account personalized generation requirements. Therefore, even if LLMs receive user-specific semantic style features, they are unable to effectively translate these features into personalized outputs. To tackle this issue, we propose a multi-user style adapter aimed at bridging the gap between generic outputs and user-specific personalized outputs, ultimately enabling personalized generation for multiple users.

Specifically, as illustrated in Fig.~\ref{fig:model}(a), at each decoding step $t$, 
the style hidden states are extracted by a simple neural network consisting of two fully connected layers as:
\begin{align}
s_t & = W_c(W_h h_t + b_h) + b_c,
\end{align}
where $s_t$ represents the style hidden states at the $t$-th step of the output sequence. $W_*$ and $b_*$ denote the trainable parameters in this section. The obtained style hidden states $s_t$ are then passed through a feed-forward layer with \texttt{Softmax} function to estimate the style probability distribution over the entire vocabulary $\mathcal{V}$ for user $u$ as:
\begin{align}
P_{\text{s}}(x_t|x_{<t},A_u;p_u) & = \texttt{Softmax}(W_s s_t + b_s),
\end{align}
where $A_u$ denotes the learned coding style attribute representations of user $u$ which appear in his code records in the training data. For simplicity, we omit the other trainable parameters. To merge the style probability distribution and the generic probability distribution of LLMs, we incorporate a dynamic gate vector $g_t$, which indicates the weight between the two distributions. 
The gate vector $g_t$ is derived by combining the hidden state 
of the decoder layer $h_t$ with the style distribution $P_{\text{s}}(x_t|x_{<t},A_u;p_u)$ as:
\begin{align}
s^{'}_t & = \texttt{Relu}(W_g P_{\text{s}}(x_t|x_{<t},A_u;p_u)+b_g), \\
h^{'}_t & = W_k h_t + b_k, \\
g_t & = \texttt{Sigmoid}(W_r(s^{'}_t + h^{'}_t) + b_r),
\end{align}
where 
$g_t \in \mathbb{R}^{|\mathcal{V}|}$ represents the dynamic gating vector for step $t$. The final probability distribution $P(x_t|x_{<t},A_u;p_u)$ is then derived as:
\begin{equation}
\begin{split}
    P(x_t|x_{<t},A_u;p_u) = g_t \cdot P_{\text{s}}(x_t|x_{<t},A_u;p_u)\\
    + (1 - g_t) \cdot P_{\text{o}}(x_t|x_{<t},A_u;p_u),
\end{split}
\end{equation}
where $P_{\text{o}} \in \mathbb{R}^{|\mathcal{V}|} $ denotes the generic distribution without the style adapter.

\noindent\textbf{Contrastive Learning.} In our approach, the use of a shared style adapter by multiple users necessitates a clear distinction in style hidden states among users.  
Therefore, we incorporate a contrastive learning strategy into our model to aid in learning style hidden states based on global style features, which includes both syntax and semantic style features. 
Specifically, we define the global style features $\hat{p}$ and global style hidden states $\hat{s}$ respectively as:
\begin{align}
\hat{p} &= \frac{1}{m} \sum_{i=1}^{m} p_i + \frac{1}{k} \sum_{i=1}^{k} a_i,\\
\hat{s}&= \frac{1}{m+n} \sum_{t=1}^{m+n} s_t. 
\end{align}
We aim to maximize
the correlation between $\hat{p}$ and $\hat{s}$ of the same user while minimize the correlation with other non-matching
pairs. 
Specifically, we formulate the contrastive objective as:
\begin{align}
\mathcal{L}_{\text{CL}}= -\log \frac{\exp (corr(\hat{p}_u , \hat{s}_u) / \tau)}{\sum_{u^-} \exp (corr(\hat{p}_u, \hat{s}_{u^-}) / \tau)},
\end{align}

where $\tau$ is the temperature parameter. By minimizing this loss, the style hidden states can be optimized to enhance personalized expressiveness.
\noindent\textbf{Training Objective.} In this training stage, the generation loss is defined as:
\begin{align}
\mathcal{L}_{\text{imp}}= -\sum_{t=1}^{n} \log P(x_t|x_{<t},A_u;p_u).
\vspace{-0.5em}
\end{align}
We train implicit style features, the output layer, and the multi-user style adapter with the contrastive loss jointly as:
\begin{align}
\mathcal{L}=\mathcal{L}_{\text{imp}}+\alpha\mathcal{L}_{\text{CL}},
\end{align}
where \(\alpha\) is a hyper-parameter.

\noindent

\subsection{Inference}
\label{sec:Two-train}
 During the inference stage, when presented with a programming question $q$ for a specific user $u$, we are able to gather the coding style attributes $A_u$ and implicit style features $p_u$. We then utilize the same prompt template as the second training stage to generate personalized code $c$ for user $u$.

\section{Experiments}
\subsection{Experimental Setup}
\paragraph{Dataset.} 
Until now, no public dataset is available for personalized code generation task. 
In this study, we first build two datasets for this novel task.  
Our datasets are collected from PTA\footnote{https://pintia.cn/}(Programming Teaching Assistant), 
for enhancing students' programming skills and will be made public upon acceptance of this work. 
The platform has recorded different users' problem-solving histories for different programming problems. 
For each user, we can collect all his/her written code for solving different problems. 
After our empirical exploration, most platform users have a few problem-solving records, while only a small subset have extensive records. 

Thus we construct two datasets \textbf{PCISparse} 
 (Personalized Code Interaction Sparse) and \textbf{PCIDense} (Personalized Code Interaction Dense), based on problem sets with average user interaction records below and above 100. 
Overall statistics of the two dataset are given in Table \ref{citation1-guide}.
After data collection, we split the dataset into training, validation and testing set by the ratio of 8:1:1. 
Further details can be found in Appendix~\ref{subsec:A1}.

\paragraph{Baselines.} We compare our model with state-of-the-arts methods as follows: (1) \textbf{CodeLlama}: CodeLlama-Instruct-7B~\cite{roziere2023code} can be seen as a benchmark  without personalization; (2) \textbf{DAPT}: Domain Adaptive Pre-Training~\cite{gururangan2020don}; (3) \textbf{L-LDB}: Customization for a specific software project~\cite{zlotchevski2022exploring}; (4) \textbf{Adapter}: Adapter is used to fine-tune the LLM~\cite{houlsby2019parameter}.Further details can be found in Appendix~\ref{subsec:A2}.

\begin{table}[t]
\small
\resizebox{0.48\textwidth}{!}{
\begin{tabular}{lccccc}
\toprule 
\multirow{2}{*}{\textbf{Dataset}} & \multirow{2}{*}{\textbf{records}} & \multirow{2}{*}{\textbf{users} } & \multicolumn{3}{c}{\textbf{records per user} } \\
& & & \textbf{max} & \textbf{min} & \textbf{avg}\\
\midrule 
PCIDense & 5794 & 50 & 178 & 101 & 115\\
PCISparse & 34642 & 1121 & 113 & 17 & 31\\
\bottomrule
\end{tabular}
}
\vspace{-0.5em}
\caption{Dataset Statistics.}
\label{citation1-guide}
\vspace{-1em}
\end{table}

\paragraph{Implementation Details.}

We choose CodeLlama~\cite{roziere2023code} as the base LLM. For training, the model is optimized with AdamW~\cite{loshchilov2017decoupled} and Fully Sharded Data Parallel~\cite{ott2021fully}. 
During decoding, code is generated using greedy decoding. Further details can be found in Appendix~\ref{subsec:A3}.

\begin{table*}[!thb]
\resizebox{\textwidth}{!}{
\begin{tabular}{llllll|llll}
\hline 
\multirow{2}{*}{ Number of users }&\multirow{2}{*}{ Model } & \multicolumn{4}{c}{ PCIDense } & \multicolumn{4}{c}{ PCISparse }\\
&& CSS & BLEU  & Rouge-1 & Rouge-2 & CSS & BLEU  & Rouge-1 & Rouge-2\\
\hline 
\multirow{4}{*}{ Single } & CodeLlama & 48.73&48.09&39.23&25.11&46.29&50.28&37.49&24.20
\\
&DAPT & 27.38 &30.78&40.01&28.77&34.29&39.21&49.17&38.30\\
&L-LDB & \textbf{64.06} &54.93&46.35&34.12&\textbf{61.09}&57.89&44.18&31.53\\
&Adapter	&49.44&49.72&43.95&31.12&39.78&42.16&25.16&13.84\\
\hline
\multirow{4}{*}{ Multiple }&\textbf{\sc MPCoder}\textsubscript{ISF}& 56.68 &56.35&42.67&29.06&60.34&58.10&41.53&27.56 \\
&\textbf{\sc MPCoder}\textsubscript{EFS}& 57.22&56.17&41.63&27.02&62.52&55.71&40.50&26.73 \\
&\textbf{\sc MPCoder}\textsubscript{IES}& 63.26 &55.72&42.73&29.17&65.61&56.67&40.91&27.26 \\
&\textbf{\sc MPCoder}& \textbf{64.50}&55.73&41.56&28.25&\textbf{66.18}&57.50&41.74&28.11 \\
\hline
\end{tabular}
}
\caption{\label{citation2-guide}
Evaluation results on the Java personalized code generation dataset PCIDense and PCISparse. All results in the table are reported in percentage (\(\%\)). ``Single'' represents the model can only be used for a single user. We report the BLEU and Rouge scores for reference by calculating BLEU-4 and Rouge-1/2.
}
\vspace{-1em}
\end{table*}

\subsection{Evaluation Metrics}
\paragraph{CSS Evaluation Metrics.} 

The personalized code generation task is more concerned with generated code styles, the traditional evaluation metrics such as BLEU~\cite{papineni2002bleu} and Rouge~\cite{lin2004automatic} scores neglect important syntactic and semantic features of code and are not suitable for evaluating coding styles. 

Inspired by the empirical research on coding styles~\cite{zou2019does}, we first propose an evaluation metric, namely Coding Style Score (CSS), for evaluating coding styles between different codes. 

Specifically, we characterized the coding style with 24 style criteria (detail in Appendix~\ref{subsec:B2}) for Java programming language. 
These criteria are associated with three aspects of coding styles: code structure, formatting, and naming. 
These style criteria also comply with Google's Java coding style\footnote{http://google.github.io/styleguide/javaguide.html}. 
For a given Java code file, it can be parsed with a 24-dimensional coding style criteria vector. 
Each dimension signifies the percentage of a specific criteria violation.

More formally, the coding style criteria vector of a Java file could be defined as $c = \langle c_{1}, c_{2}, ..., c_{24}\rangle$, 
where \(c_i \in [0,1]\) describes the extent of the $i$-th coding style criteria has been violated. 
We define the CSS metric between the generated code \(c_{\text{gen}}\) and the reference code \(c_{\text{ref}}\) as follows:
\begin{align}
\text{css}(c_{\text{gen}}, c_{\text{ref}}) = 1 - D_{JS}(c_{\text{gen}}, c_{\text{ref}}),
\end{align}
where \(\text{css} \in [0,1]\), and \(D_{JS}\) is JS (Jensen-Shannon) divergence: it measures the similarity of two probability distributions as:
\begin{align}
D_{JS}(p, q) =\frac{1}{2}[D_{kl}(p \| \frac{p+q}{2}) +D_{kl}(q \| \frac{p+q}{2})],
\end{align}
where \(D_{kl}\) is Kullback-Leibler Divergence as:
\begin{align}
D_{kl}(p\|q) &=\sum_{i=1}^n p(x_i) \log(\frac{p(x_i)}{q(x_i)}).
\end{align}
With CSS evaluation metric, we can provide a quantitative way to measure the coding style between different codes. 
The larger CSS metric is, the more similar the coding styles of the generated code and the reference code are.

\paragraph{Correctness Evaluation Metrics.}
 
Regarding our task, we hope to not only generate personalized code, but also maintaining the correctness of the generated code at the same time. 
We used HumanEval-X\footnote{https://github.com/THUDM/CodeGeeX2} to evaluate the correctness of the generated code.
HumanEval-X contains 164 Java programming problems and their corresponding test cases, a  code is considered as correct if it passes all test cases for a specific programming problem. 

\begin{table}[!htb]
\resizebox{0.48\textwidth}{!}{
\sisetup{output-exponent-marker=\ensuremath{\mathrm{e}},exponent-product=\times}
\begin{tabular}{lccc}
\hline
\multirow{2}{*}{ Model } & \multirow{2}{*}{ multi-user }& \multicolumn{2}{c}{ Parameters }\\
& & Total & Trained\\
\hline
DAPT &\textcolor{red}{\XSolidBrush} &6.7B*N&100\%*N\\
L-LDB &\textcolor{red}{\XSolidBrush} &6.7B*N& 3.00\%*N\\
Adapter &\textcolor{red}{\XSolidBrush} & (6.7B+1.2M)*N & 0.02\%*N\\
\hline
\textbf{\sc MPCoder}\textsubscript{ISF}& \textcolor{green}{\checkmark}&20K*N+6.7B & \num{3e-4}\% * N + 1.95\%\\
\textbf{\sc MPCoder}\textsubscript{ESF}& \textcolor{green}{\checkmark}&96K+6.7B & 1.95\%\\
\textbf{\sc MPCoder}\textsubscript{IES}	& \textcolor{green}{\checkmark}&20K*N+96K+6.7B & \num{3e-4}\% * N+1.95\% \\
\textbf{\sc MPCoder}	&\textcolor{green}{\checkmark}& 20K*N+96K+6.9B & \num{3e-4}\% * N+\num{4.82}\%\\
\hline
\end{tabular}
}
\caption{\label{citation3-guide}
Comparison of the total size and trainable parameters. N denotes the number of users and `multi-user' indicates whether the model supports multi-user. 96K is the number of parameters for 25 explicit coding style attributes. 
}
\vspace{-1.5em}
\end{table}
\subsection{Experimental Results}
\paragraph{Coding Style Evaluation.}  
Table~\ref{citation2-guide} shows the results of different models on PCIDense and PCISparse datasets respectively. 
It is obvious that: (1) {\sc MPCoder} outperforms all other baselines in terms of the CSS evaluation metric, which verifies the advantage of our approach in learning code styles from code. 
(2) We include three variants of {\sc MPCoder}, namely {\sc MPCode}\textsubscript{ISF}, {\sc MPCode}\textsubscript{ESF} and {\sc MPCoder}\textsubscript{IES}, as baselines. 
The {\sc MPCoder}\textsubscript{ISF} only uses the implicit style features as input; The {\sc MPCoder}\textsubscript{ESF} only uses the explicit style features as input; The {\sc MPCoder}\textsubscript{IES} uses the implicit and explicit style features without changing the structure of the CodeLlama (\textit{i.e.}, not adding Multi-user Style Adapter and Contrastive Learning). The experimental results show that incorporating implicit features can notably enhance CodeLlama's performance in terms of textual similarities (\textit{i.e.}, BLEU and Rouge) and coding style similarities (\textit{i.e.}, CSS). 
After adding explicit features ({\sc MPCoder}\textsubscript{IES}), the textual similarities almost stay the same while the CSS has been further significantly improved. 
This confirms the effectiveness of incorporating explicit and implicit features learning in our study. 
(3) The L-LDB is a customized baseline tailored to individual users, which has its advantage compared with other baselines. However, as it is specifically designed for single users, it requires the model to be retrained for each user, making it cost-prohibitive for multi-user code generation.  Qualitative examples of our model and other models can be found in Appendix~\ref{subsec:B4}.

\noindent\textbf{Cost Analysis.} 
Table~\ref{citation3-guide} shows the total size and trainable parameters of each model. 
The results show that {\sc MPCoder} can greatly reduce training and storage costs for multiple users. 
The cost of adding new users to our model can be neglected ({3e-4}\% in parameters per user).

\begin{figure}[!thb]
\small
\centering
\includegraphics[width=0.40\textwidth]{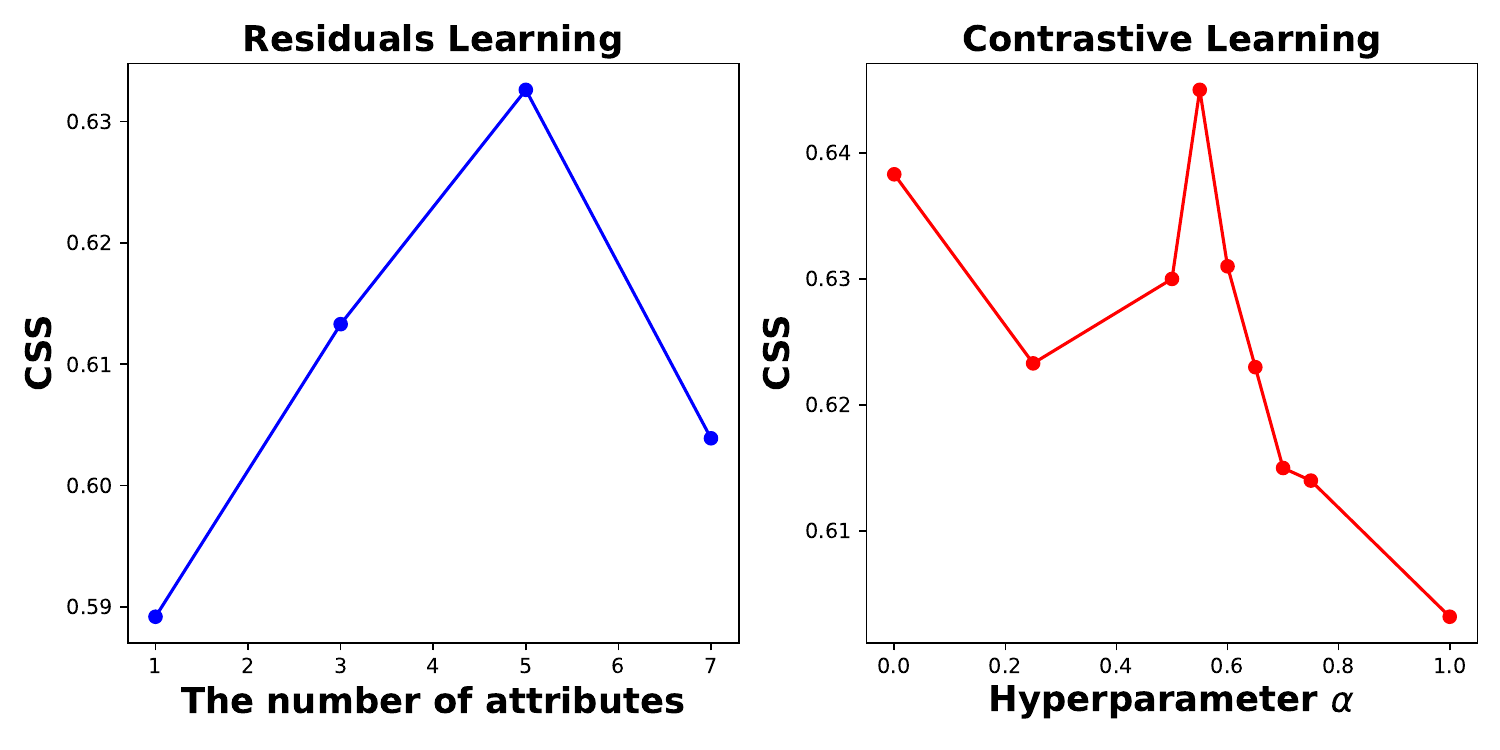}
\vspace{-1em}
\caption{Residual Learning~(left) and Contrastive Learning Hyperparameter~(right) Effects on CSS.}

\label{fig:Residual}
\vspace{-1em}
\end{figure}

\begin{figure}[!thb]
\small
\centering
\includegraphics[width=0.45\textwidth]{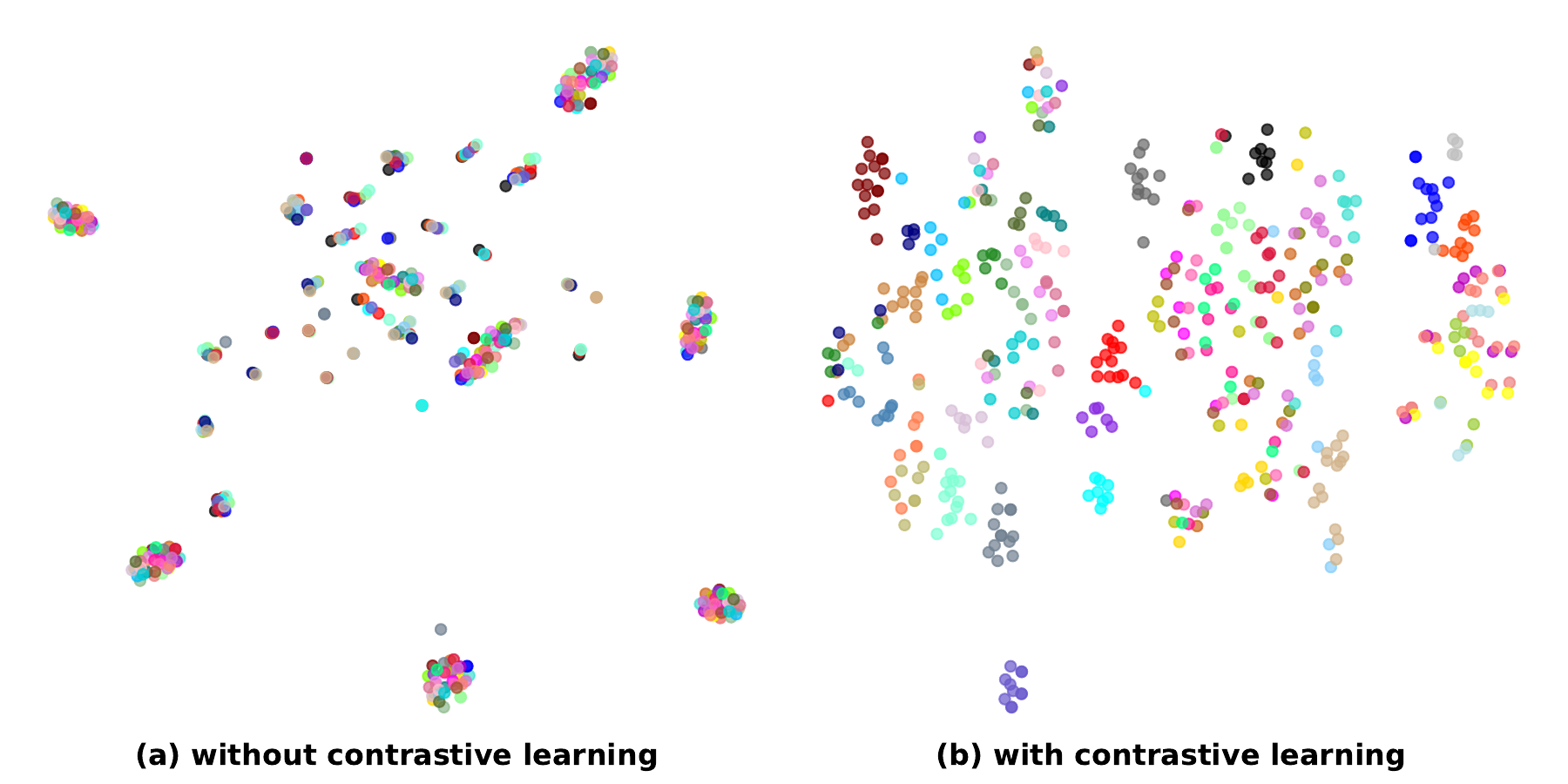}
\vspace{-0.5em}
\caption{t-SNE visualization results: style hidden states of user interaction records on PCIDense. Different colors denote different users.
}

\label{fig:contrastive-Analysis}
\vspace{-2em}
\end{figure}
\paragraph{Residual and Contrastive Learning Analysis.}
There are two key hyperparameters, which are the number of attributes in residual learning and the weight parameter $\alpha$ in the loss function. 
In this section, we evaluate the the influence of these two parameters. 

The number of attributes means the maximum number of style attributes in the attribute-residual prompt. 
As shown in Fig.~\ref{fig:Residual}(a), when setting the number of attributes to 1, it equals to not utilizing residual learning. 
The experimental results show that employing residual learning with 5 attributes can effectively guide LLM to understand code attributes.

For hyperparameters \(\alpha\), as shown in Fig.~\ref{fig:Residual}(b), we vary its value from 0 to 1.0 with a step size of 0.25. Then we further adjust it from 0.50 to 0.75 with a step size of 0.05. The experimental results show that setting \(\alpha\) to 0.55 can achieve optimal performance.

Furthermore, we visually represent the style hidden states of 50 users to demonstrate the effectiveness of contrastive learning in Fig.~\ref{fig:contrastive-Analysis}. The hidden states of records from the same user are clustered together, while those from different users are effectively separated.

\begin{table}[t]
\centering
\small
\resizebox{0.48\textwidth}{!}{
\begin{tabular}{lll}
\hline 
\multirow{2}{*}{ Model } & \multicolumn{2}{c}{ CSS } \\
& PCIDense & PCISparse \\
\hline
\textbf{\sc MPCoder}& \textbf{64.64} &\textbf{66.18} \\
w/o Coding Style Attributes&58.19&64.13 \\
w/o Contrastive Learning&63.83&63.36 \\
w/o Multi-user Style Adapter&63.26&65.61 \\
\hline
\end{tabular}
}
\vspace{-0.5em}
\caption{\label{citation4-guide}
Ablation study.
}
\vspace{-1em}
\end{table}

\paragraph{Ablation Study.} As illustrated in Table~\ref{citation4-guide}, we performed an ablation study by removing key components (\textit{i.e.}, Coding Style Attributes, Contrastive Learning and Muti-user Style Adapter) of {\sc MPCoder} separately. 
The experimental results show that: (1) No matter which component we drop, it hurts the overall performance of our model, which signals the importance and effectiveness of all three components.
(2) The CSS drops most significantly on PCIDense and PCISparse datasets when the Coding Style Attributes and Contrastive Learning component are removed, showing these two components can complement each other under different situations, which justifies the importance and
necessity of the above two components.

\paragraph{Correctness Evaluation.}  
We aim to generate personalized code while still keeping the code correct. We further evaluate the correctness of each baseline method on HumanEval-X dataset. 
We report the average accuracy of the generated code based on three prompts. 
Further details can be found in Appendix~\ref{subsec:B1}. 
As shown in Table~\ref{citation5-guide}, DAPT is unable to generate compilable code due to an excessive focus on word overlap, resulting in repeated instances such as `doublee' for data types. 
From the table, we can observe that:
(1) Although L-LDB performs relatively well in personalized code generation, its code correctness has been greatly affected compared with CodeLlama.
(2) Adapter maintains the generated code correctness with CodeLlama. However, it is not suitable for generating personalized code. The code generated by {\sc MPCoder} achieves a good balance between correctness and personalization compared to all baselines. 
\begin{table}
\centering
\resizebox{0.48\textwidth}{!}{
\begin{tabular}{clcc}
\hline
\multirow{1}{*}{Numbers of user} &\multirow{1}{*}{ Model } & \multicolumn{1}{c}{ PCIDense} & \multicolumn{1}{c}{PCISparse}\\
\hline
\multirow{4}{*}{ Single } &CodeLlama  & 30.49\% & 30.49\%\\
&DAPT & - & -\\
&L-LDB & \textbf{28.62}\%& 22.56\%\\
&Adapter	 & 27.32\% & \textbf{29.73}\%\\
\hline
\multirow{4}{*}{Multiple} &\textbf{\sc MPCoder}\textsubscript{ISF}&  29.57\% & 25.76\%\\
&\textbf{\sc MPCoder}\textsubscript{ESF}	 & 30.24\% & \textbf{28.53}\%\\
&\textbf{\sc MPCoder}\textsubscript{IES}	 & 27.88\% & 26.60\%\\

&\textbf{\sc MPCoder}	 & \textbf{31.25}\%& 26.18\%\\
\hline
\end{tabular}
}
\vspace{-0.5em}
\caption{\label{citation5-guide}
Correctness evaluation.
}
\vspace{-1em}
\end{table}

\paragraph{Human Study.} 
To verify the effectiveness of our CSS metric, we conduct a human study to compare the results of CSS with human results. In particular, we compare {\sc MPCoder} with L-LDB and Adapter by conducting a user study on PCIDense. We asked 5 users to answer questionnaires of 60 comparative questions, totaling 300 answers. All questions present them with a choice between two options. Users are asked to answer the question: ``Which of the two code copies is closer in coding style to the reference code?'', and every user is provided with three options (i.e., A is Better, B is Better, Cannot Determine/Both Equally). Further detail can be found in Appendix~\ref{subsec:A4}. We calculate the CSS values of the two codes and the reference code for different models respectively. The model with the highest CSS value is regarded as the best model. Table~\ref{citation6-guide} shows the results of the human study.
\begin{table}
\centering
\resizebox{0.48\textwidth}{!}{
\begin{tabular}{clcc}
\hline
\multirow{2}{*}{Numbers of user} &\multirow{2}{*}{ Model } & \multicolumn{2}{c}{ Rate of the best models}\\
&&\multicolumn{1}{c}{ human evaluation} & \multicolumn{1}{c}{ CSS}\\
\hline
 Single &L-LDB  & 37\% & 41\%\\
Single &Adapter & 12\% & 13\%\\
Multiple &\textbf{\sc MPCoder}&  43\% & 46\%\\
-&Undecided&8\%&-\\
\hline
\end{tabular}
}
\vspace{-0.5em}
\caption{\label{citation6-guide}
Human study.
}
\vspace{-1em}
\end{table}

The results obtained by CSS are consistent with human evaluation({\sc MPCoder} > L-LDB > Adapter), which verifies the effectiveness of our proposed CSS evaluation metric for estimating coding styles. By comparing the results of different models, {\sc MPCoder} outperforms the L-LDB and Adapter significantly and consistently. The experimental results show our model's superiority in both automatic evaluation (CSS) and human evaluation.

\paragraph{Adaptation For New Users.} 
Introducing a new user to our model can be categorized into two scenarios: (1) the new user's historical coding records are available; (2) the new user's historical coding records are unavailable. In Appendix~\ref{subsec:A5}, we discuss in detail how {\sc MPCoder} effectively adapts to these two types of new users.  
\section{Related Work}
\paragraph{Code Pre-trained Language Models.} With the latest developments in the Transformer-based model~\cite{vaswani2017attention},  recent work has attempted to apply LLM to code to advance software engineering and code intelligence.
CodeBERT~\cite{feng2020codebert} pretrains the NL-PL data. CodeT5~\cite{wang2021codet5} leverages the T5~\cite{raffel2020exploring} architecture to leverage code semantics through identifier tokens. 
LLM~\cite{radford2019language,brown2020language} has made many achievements in the field of natural language processing. the OpenAI Codex~\cite{chen2021evaluating} model with 12B parameters pioneered and demonstrated the potential of large code generation models pre-trained on billions of lines of public code. Subsequently, models dedicated to code generation emerged, such as CodeLlama~\cite{roziere2023code} and CodeGeex~\cite{zheng2023codegeex}.
\paragraph{Personalized Generation.} Most of the existing work in personalized generation focuses on attribute-based controlled text generation~\cite{keskar2019ctrl}, such as emotions and topics~\cite{dathathri2019plug,kong2021stylized,xu2022long}. The text-description-based approach~\cite{song2021bob} focuses on promoting character consistency through pre-trained language models. The embedding-based utilizes user ID information~\cite{al2016conversational} or embedded user dialogue~\cite{ma2021one} history as an implicit profile.
Within the domain of personalized code generation, existing approaches ~\cite{zlotchevski2022exploring} involves fine-tuning for specific software projects, providing Java unit tests for a single coding style, while we focus on providing coding styles for multiple users.

\section{Conclusions}
This research aims to generate personalized code for multiple users to satisfy the coding styles of different developers or projects. 
To perform this novel task, we propose an approach {\sc MPCoder} which utilizes explicit coding style residual learning to capture the syntax style standards and implicit coding style learning to capture the semantic style of each user. 
We propose a multi-user style adapter to bridge the gap between the generic outputs and user-specific personalized outputs. 
{\sc MPCoder} can ultimately generate personalized code for multiple users simultaneously. 
The experimental results show the effectiveness of our for this task. 
We hope our study can lay the foundations for this new research and provide valuable insights into the potential for personalized generation capabilities of general LLMs.

\section{Limitations}
Several limitations are concerned with our work. 
Firstly, our study is based on Java, which is one of the most popular programming languages used by developers. 
However, our approach is language-independent, we believe our approach can be easily adapted to other programming languages such as Python or Javascript.

Secondly, the correctness of the generated code has been affected when our model was applied to the PCISparse dataset. 
Exploring effective ways to generate personalized code while maintaining its correctness with a limited number of data samples is an interesting research topic for our future work.

\section{Ethics Statement}
To prevent privacy leaks, we have removed personal and sensitive information from our dataset, utilizing anonymous IDs as individual identifiers. Specifically, the raw data underwent an initial preprocessing step to transform it into structured data. We manually identified labels that may pose privacy risks (e.g., ID, user name, email address, age, gender), and then anonymized the corresponding information by either deleting it or mapping it to new values.
We explore the feasibility of using LLMs to perform personalized code generation. 
However, LLMs such as CodeLlama may have some ethical biases, and these ethical concerns inevitably affect our proposed approach Ethical guidelines and the deployment of such techniques should be considered to mitigate potential negative consequences. 
We hope our work will stimulate further investigation and advancement in this novel research area of personalized code generation and the general personalized generation abilities of LLMs. 

\section{Acknowledgement}
This work was supported by the National Natural Science Foundation of China (No.62037001, No.62307032, No.62293555) and the Shanghai Rising-Star Program (23QA1409000).

\bibliography{acl24}

\begin{thebibliography}{45}
\expandafter\ifx\csname natexlab\endcsname\relax\def\natexlab#1{#1}\fi

\bibitem[{Ahmad et~al.(2021)Ahmad, Chakraborty, Ray, and Chang}]{ahmad2021unified}
Wasi~Uddin Ahmad, Saikat Chakraborty, Baishakhi Ray, and Kai-Wei Chang. 2021.
\newblock Unified pre-training for program understanding and generation.
\newblock \emph{arXiv preprint arXiv:2103.06333}.

\bibitem[{Al-Rfou et~al.(2016)Al-Rfou, Pickett, Snaider, Sung, Strope, and Kurzweil}]{al2016conversational}
Rami Al-Rfou, Marc Pickett, Javier Snaider, Yun-Hsuan Sung, Brian Strope, and Ray Kurzweil. 2016.
\newblock Conversational contextual cues: The case of personalization and history for response ranking.
\newblock \emph{arXiv preprint arXiv:1606.00372}.

\bibitem[{Alayrac et~al.(2022)Alayrac, Donahue, Luc, Miech, Barr, Hasson, Lenc, Mensch, Millican, Reynolds et~al.}]{alayrac2022flamingo}
Jean-Baptiste Alayrac, Jeff Donahue, Pauline Luc, Antoine Miech, Iain Barr, Yana Hasson, Karel Lenc, Arthur Mensch, Katherine Millican, Malcolm Reynolds, et~al. 2022.
\newblock Flamingo: a visual language model for few-shot learning.
\newblock \emph{Advances in Neural Information Processing Systems}, 35:23716--23736.

\bibitem[{Alkhatib(1992)}]{alkhatib1992maintenance}
Ghazi Alkhatib. 1992.
\newblock The maintenance problem of application software: An empirical analysis.
\newblock \emph{Journal of Software Maintenance: Research and Practice}, 4(2):83--104.

\bibitem[{Allamanis et~al.(2014)Allamanis, Barr, Bird, and Sutton}]{allamanis2014learning}
Miltiadis Allamanis, Earl~T Barr, Christian Bird, and Charles Sutton. 2014.
\newblock Learning natural coding conventions.
\newblock In \emph{Proceedings of the 22nd acm sigsoft international symposium on foundations of software engineering}, pages 281--293.

\bibitem[{Brown et~al.(2020)Brown, Mann, Ryder, Subbiah, Kaplan, Dhariwal, Neelakantan, Shyam, Sastry, Askell et~al.}]{brown2020language}
Tom Brown, Benjamin Mann, Nick Ryder, Melanie Subbiah, Jared~D Kaplan, Prafulla Dhariwal, Arvind Neelakantan, Pranav Shyam, Girish Sastry, Amanda Askell, et~al. 2020.
\newblock Language models are few-shot learners.
\newblock \emph{Advances in neural information processing systems}, 33:1877--1901.

\bibitem[{Chen et~al.(2021)Chen, Tworek, Jun, Yuan, Pinto, Kaplan, Edwards, Burda, Joseph, Brockman et~al.}]{chen2021evaluating}
Mark Chen, Jerry Tworek, Heewoo Jun, Qiming Yuan, Henrique Ponde de~Oliveira Pinto, Jared Kaplan, Harri Edwards, Yuri Burda, Nicholas Joseph, Greg Brockman, et~al. 2021.
\newblock Evaluating large language models trained on code.
\newblock \emph{arXiv preprint arXiv:2107.03374}.

\bibitem[{Chen et~al.(2022)Chen, Sun, Zhu, Li, Lu, and Gao}]{chen2022cat}
Nuo Chen, Qiushi Sun, Renyu Zhu, Xiang Li, Xuesong Lu, and Ming Gao. 2022.
\newblock Cat-probing: A metric-based approach to interpret how pre-trained models for programming language attend code structure.
\newblock \emph{arXiv preprint arXiv:2210.04633}.

\bibitem[{Cheng et~al.(2022)Cheng, Murphy-Hill, Canning, Jaspan, Green, Knight, Zhang, and Kammer}]{cheng2022improves}
Lan Cheng, Emerson Murphy-Hill, Mark Canning, Ciera Jaspan, Collin Green, Andrea Knight, Nan Zhang, and Elizabeth Kammer. 2022.
\newblock What improves developer productivity at google? code quality.
\newblock In \emph{Proceedings of the 30th ACM Joint European Software Engineering Conference and Symposium on the Foundations of Software Engineering}, pages 1302--1313.

\bibitem[{Dathathri et~al.(2019)Dathathri, Madotto, Lan, Hung, Frank, Molino, Yosinski, and Liu}]{dathathri2019plug}
Sumanth Dathathri, Andrea Madotto, Janice Lan, Jane Hung, Eric Frank, Piero Molino, Jason Yosinski, and Rosanne Liu. 2019.
\newblock Plug and play language models: A simple approach to controlled text generation.
\newblock \emph{arXiv preprint arXiv:1912.02164}.

\bibitem[{Feng et~al.(2020)Feng, Guo, Tang, Duan, Feng, Gong, Shou, Qin, Liu, Jiang et~al.}]{feng2020codebert}
Zhangyin Feng, Daya Guo, Duyu Tang, Nan Duan, Xiaocheng Feng, Ming Gong, Linjun Shou, Bing Qin, Ting Liu, Daxin Jiang, et~al. 2020.
\newblock Codebert: A pre-trained model for programming and natural languages.
\newblock \emph{arXiv preprint arXiv:2002.08155}.

\bibitem[{Guo et~al.(2021)Guo, Tan, Liu, Xing, and Hu}]{guo2021text}
Han Guo, Bowen Tan, Zhengzhong Liu, Eric~P Xing, and Zhiting Hu. 2021.
\newblock Text generation with efficient (soft) q-learning.
\newblock \emph{arXiv e-prints}, pages arXiv--2106.

\bibitem[{Gururangan et~al.(2020)Gururangan, Marasovi{\'c}, Swayamdipta, Lo, Beltagy, Downey, and Smith}]{gururangan2020don}
Suchin Gururangan, Ana Marasovi{\'c}, Swabha Swayamdipta, Kyle Lo, Iz~Beltagy, Doug Downey, and Noah~A Smith. 2020.
\newblock Don't stop pretraining: Adapt language models to domains and tasks.
\newblock \emph{arXiv preprint arXiv:2004.10964}.

\bibitem[{Houlsby et~al.(2019)Houlsby, Giurgiu, Jastrzebski, Morrone, De~Laroussilhe, Gesmundo, Attariyan, and Gelly}]{houlsby2019parameter}
Neil Houlsby, Andrei Giurgiu, Stanislaw Jastrzebski, Bruna Morrone, Quentin De~Laroussilhe, Andrea Gesmundo, Mona Attariyan, and Sylvain Gelly. 2019.
\newblock Parameter-efficient transfer learning for nlp.
\newblock In \emph{International Conference on Machine Learning}, pages 2790--2799. PMLR.

\bibitem[{Hu et~al.(2021)Hu, Shen, Wallis, Allen-Zhu, Li, Wang, Wang, and Chen}]{hu2021lora}
Edward~J Hu, Yelong Shen, Phillip Wallis, Zeyuan Allen-Zhu, Yuanzhi Li, Shean Wang, Lu~Wang, and Weizhu Chen. 2021.
\newblock Lora: Low-rank adaptation of large language models.
\newblock \emph{arXiv preprint arXiv:2106.09685}.

\bibitem[{Husain et~al.(2019)Husain, Wu, Gazit, Allamanis, and Brockschmidt}]{husain2019codesearchnet}
Hamel Husain, Ho-Hsiang Wu, Tiferet Gazit, Miltiadis Allamanis, and Marc Brockschmidt. 2019.
\newblock Codesearchnet challenge: Evaluating the state of semantic code search.
\newblock \emph{arXiv preprint arXiv:1909.09436}.

\bibitem[{Keskar et~al.(2019)Keskar, McCann, Varshney, Xiong, and Socher}]{keskar2019ctrl}
Nitish~Shirish Keskar, Bryan McCann, Lav~R Varshney, Caiming Xiong, and Richard Socher. 2019.
\newblock Ctrl: A conditional transformer language model for controllable generation.
\newblock \emph{arXiv preprint arXiv:1909.05858}.

\bibitem[{Kong et~al.(2021)Kong, Huang, Tung, Guan, and Huang}]{kong2021stylized}
Xiangzhe Kong, Jialiang Huang, Ziquan Tung, Jian Guan, and Minlie Huang. 2021.
\newblock Stylized story generation with style-guided planning.
\newblock \emph{arXiv preprint arXiv:2105.08625}.

\bibitem[{Kropp and Meier(2013)}]{kropp2013teaching}
Martin Kropp and Andreas Meier. 2013.
\newblock Teaching agile software development at university level: Values, management, and craftsmanship.
\newblock In \emph{2013 26th International Conference on Software Engineering Education and Training (CSEE\&T)}, pages 179--188. IEEE.

\bibitem[{Lester et~al.(2021)Lester, Al-Rfou, and Constant}]{lester2021power}
Brian Lester, Rami Al-Rfou, and Noah Constant. 2021.
\newblock The power of scale for parameter-efficient prompt tuning.
\newblock \emph{arXiv preprint arXiv:2104.08691}.

\bibitem[{Li et~al.(2023{\natexlab{a}})Li, Zhang, and Chen}]{li2023personalized}
Lei Li, Yongfeng Zhang, and Li~Chen. 2023{\natexlab{a}}.
\newblock Personalized prompt learning for explainable recommendation.
\newblock \emph{ACM Transactions on Information Systems}, 41(4):1--26.

\bibitem[{Li et~al.(2023{\natexlab{b}})Li, Allal, Zi, Muennighoff, Kocetkov, Mou, Marone, Akiki, Li, Chim et~al.}]{li2023starcoder}
Raymond Li, Loubna~Ben Allal, Yangtian Zi, Niklas Muennighoff, Denis Kocetkov, Chenghao Mou, Marc Marone, Christopher Akiki, Jia Li, Jenny Chim, et~al. 2023{\natexlab{b}}.
\newblock Starcoder: may the source be with you!
\newblock \emph{arXiv preprint arXiv:2305.06161}.

\bibitem[{Li et~al.(2022{\natexlab{a}})Li, Choi, Chung, Kushman, Schrittwieser, Leblond, Eccles, Keeling, Gimeno, Dal~Lago et~al.}]{li2022competition}
Yujia Li, David Choi, Junyoung Chung, Nate Kushman, Julian Schrittwieser, R{\'e}mi Leblond, Tom Eccles, James Keeling, Felix Gimeno, Agustin Dal~Lago, et~al. 2022{\natexlab{a}}.
\newblock Competition-level code generation with alphacode.
\newblock \emph{Science}, 378(6624):1092--1097.

\bibitem[{Li et~al.(2022{\natexlab{b}})Li, Chen, Chen, Zou, and Xu}]{li2022ropgen}
Zhen Li, Guenevere Chen, Chen Chen, Yayi Zou, and Shouhuai Xu. 2022{\natexlab{b}}.
\newblock Ropgen: Towards robust code authorship attribution via automatic coding style transformation.
\newblock In \emph{Proceedings of the 44th International Conference on Software Engineering}, pages 1906--1918.

\bibitem[{Lin and Och(2004)}]{lin2004automatic}
Chin-Yew Lin and Franz~Josef Och. 2004.
\newblock Automatic evaluation of machine translation quality using longest common subsequence and skip-bigram statistics.
\newblock In \emph{Proceedings of the 42nd Annual Meeting of the Association for Computational Linguistics (ACL-04)}, pages 605--612.

\bibitem[{Loshchilov and Hutter(2017)}]{loshchilov2017decoupled}
Ilya Loshchilov and Frank Hutter. 2017.
\newblock Decoupled weight decay regularization.
\newblock \emph{arXiv preprint arXiv:1711.05101}.

\bibitem[{Lu et~al.(2021)Lu, Guo, Ren, Huang, Svyatkovskiy, Blanco, Clement, Drain, Jiang, Tang et~al.}]{lu2021codexglue}
Shuai Lu, Daya Guo, Shuo Ren, Junjie Huang, Alexey Svyatkovskiy, Ambrosio Blanco, Colin Clement, Dawn Drain, Daxin Jiang, Duyu Tang, et~al. 2021.
\newblock Codexglue: A machine learning benchmark dataset for code understanding and generation.
\newblock \emph{arXiv preprint arXiv:2102.04664}.

\bibitem[{Ma et~al.(2021)Ma, Dou, Zhu, Zhong, and Wen}]{ma2021one}
Zhengyi Ma, Zhicheng Dou, Yutao Zhu, Hanxun Zhong, and Ji-Rong Wen. 2021.
\newblock One chatbot per person: Creating personalized chatbots based on implicit user profiles.
\newblock In \emph{Proceedings of the 44th international ACM SIGIR conference on research and development in information retrieval}, pages 555--564.

\bibitem[{Markovtsev et~al.(2019)Markovtsev, Long, Mougard, Slavnov, and Bulychev}]{markovtsev2019style}
Vadim Markovtsev, Waren Long, Hugo Mougard, Konstantin Slavnov, and Egor Bulychev. 2019.
\newblock Style-analyzer: fixing code style inconsistencies with interpretable unsupervised algorithms.
\newblock In \emph{2019 IEEE/ACM 16th International Conference on Mining Software Repositories (MSR)}, pages 468--478. IEEE.

\bibitem[{Ogura et~al.(2018)Ogura, Matsumoto, Hata, and Kusumoto}]{ogura2018bring}
Naoto Ogura, Shinsuke Matsumoto, Hideaki Hata, and Shinji Kusumoto. 2018.
\newblock Bring your own coding style.
\newblock In \emph{2018 IEEE 25th International Conference on Software Analysis, Evolution and Reengineering (SANER)}, pages 527--531. IEEE.

\bibitem[{Ott et~al.(2021)Ott, Shleifer, Xu, Goyal, Duval, and Caggiano}]{ott2021fully}
Myle Ott, Sam Shleifer, Min Xu, Priya Goyal, Quentin Duval, and Vittorio Caggiano. 2021.
\newblock Fully sharded data parallel: faster ai training with fewer gpus.

\bibitem[{Papineni et~al.(2002)Papineni, Roukos, Ward, and Zhu}]{papineni2002bleu}
Kishore Papineni, Salim Roukos, Todd Ward, and Wei-Jing Zhu. 2002.
\newblock Bleu: a method for automatic evaluation of machine translation.
\newblock In \emph{Proceedings of the 40th annual meeting of the Association for Computational Linguistics}, pages 311--318.

\bibitem[{Parr and Vinju(2016)}]{parr2016towards}
Terence Parr and Jurgen Vinju. 2016.
\newblock Towards a universal code formatter through machine learning.
\newblock In \emph{Proceedings of the 2016 ACM SIGPLAN International Conference on Software Language Engineering}, pages 137--151.

\bibitem[{Radford et~al.(2019)Radford, Wu, Child, Luan, Amodei, Sutskever et~al.}]{radford2019language}
Alec Radford, Jeffrey Wu, Rewon Child, David Luan, Dario Amodei, Ilya Sutskever, et~al. 2019.
\newblock Language models are unsupervised multitask learners.
\newblock \emph{OpenAI blog}, 1(8):9.

\bibitem[{Raffel et~al.(2020)Raffel, Shazeer, Roberts, Lee, Narang, Matena, Zhou, Li, and Liu}]{raffel2020exploring}
Colin Raffel, Noam Shazeer, Adam Roberts, Katherine Lee, Sharan Narang, Michael Matena, Yanqi Zhou, Wei Li, and Peter~J Liu. 2020.
\newblock Exploring the limits of transfer learning with a unified text-to-text transformer.
\newblock \emph{The Journal of Machine Learning Research}, 21(1):5485--5551.

\bibitem[{Roziere et~al.(2023)Roziere, Gehring, Gloeckle, Sootla, Gat, Tan, Adi, Liu, Remez, Rapin et~al.}]{roziere2023code}
Baptiste Roziere, Jonas Gehring, Fabian Gloeckle, Sten Sootla, Itai Gat, Xiaoqing~Ellen Tan, Yossi Adi, Jingyu Liu, Tal Remez, J{\'e}r{\'e}my Rapin, et~al. 2023.
\newblock Code llama: Open foundation models for code.
\newblock \emph{arXiv preprint arXiv:2308.12950}.

\bibitem[{Song et~al.(2023)Song, Jaume, Williamson, Lu, Vaidya, Miller, and Mahmood}]{song2023artificial}
Andrew~H Song, Guillaume Jaume, Drew~FK Williamson, Ming~Y Lu, Anurag Vaidya, Tiffany~R Miller, and Faisal Mahmood. 2023.
\newblock Artificial intelligence for digital and computational pathology.
\newblock \emph{Nature Reviews Bioengineering}, 1(12):930--949.

\bibitem[{Song et~al.(2021)Song, Wang, Zhang, Zhang, and Liu}]{song2021bob}
Haoyu Song, Yan Wang, Kaiyan Zhang, Wei-Nan Zhang, and Ting Liu. 2021.
\newblock Bob: Bert over bert for training persona-based dialogue models from limited personalized data.
\newblock \emph{arXiv preprint arXiv:2106.06169}.

\bibitem[{Tu et~al.(2014)Tu, Su, and Devanbu}]{tu2014localness}
Zhaopeng Tu, Zhendong Su, and Premkumar Devanbu. 2014.
\newblock On the localness of software.
\newblock In \emph{Proceedings of the 22nd ACM SIGSOFT International Symposium on Foundations of Software Engineering}, pages 269--280.

\bibitem[{Vaswani et~al.(2017)Vaswani, Shazeer, Parmar, Uszkoreit, Jones, Gomez, Kaiser, and Polosukhin}]{vaswani2017attention}
Ashish Vaswani, Noam Shazeer, Niki Parmar, Jakob Uszkoreit, Llion Jones, Aidan~N Gomez, {\L}ukasz Kaiser, and Illia Polosukhin. 2017.
\newblock Attention is all you need.
\newblock \emph{Advances in neural information processing systems}, 30.

\bibitem[{Wang et~al.(2021)Wang, Wang, Joty, and Hoi}]{wang2021codet5}
Yue Wang, Weishi Wang, Shafiq Joty, and Steven~CH Hoi. 2021.
\newblock Codet5: Identifier-aware unified pre-trained encoder-decoder models for code understanding and generation.
\newblock \emph{arXiv preprint arXiv:2109.00859}.

\bibitem[{Xu et~al.(2022)Xu, Gou, Wu, Niu, Wu, Wang, and Wang}]{xu2022long}
Xinchao Xu, Zhibin Gou, Wenquan Wu, Zheng-Yu Niu, Hua Wu, Haifeng Wang, and Shihang Wang. 2022.
\newblock Long time no see! open-domain conversation with long-term persona memory.
\newblock In \emph{Findings of the Association for Computational Linguistics: ACL 2022}, pages 2639--2650.

\bibitem[{Zheng et~al.(2023)Zheng, Xia, Zou, Dong, Wang, Xue, Wang, Shen, Wang, Li, Su, Yang, and Tang}]{zheng2023codegeex}
Qinkai Zheng, Xiao Xia, Xu~Zou, Yuxiao Dong, Shan Wang, Yufei Xue, Zihan Wang, Lei Shen, Andi Wang, Yang Li, Teng Su, Zhilin Yang, and Jie Tang. 2023.
\newblock Codegeex: A pre-trained model for code generation with multilingual evaluations on humaneval-x.
\newblock In \emph{KDD}.

\bibitem[{Zlotchevski et~al.(2022)Zlotchevski, Drain, Svyatkovskiy, Clement, Sundaresan, and Tufano}]{zlotchevski2022exploring}
Andrei Zlotchevski, Dawn Drain, Alexey Svyatkovskiy, Colin~B Clement, Neel Sundaresan, and Michele Tufano. 2022.
\newblock Exploring and evaluating personalized models for code generation.
\newblock In \emph{Proceedings of the 30th ACM Joint European Software Engineering Conference and Symposium on the Foundations of Software Engineering}, pages 1500--1508.

\bibitem[{Zou et~al.(2019)Zou, Xuan, Xie, Chen, and Xu}]{zou2019does}
Weiqin Zou, Jifeng Xuan, Xiaoyuan Xie, Zhenyu Chen, and Baowen Xu. 2019.
\newblock How does code style inconsistency affect pull request integration? an exploratory study on 117 github projects.
\newblock \emph{Empirical Software Engineering}, 24:3871--3903.

\end{thebibliography}

\appendix
\section{Experimental setup details}
\label{sec:Experimental-appendix}
\subsection{Dataset Construction Details}
\label{subsec:A1}
PCIDense(Personalized Code Interaction Dense) have 382 programming problems, PCISparse (Personalized Code Interaction Sparse) have 662 programming problems. Since codeLlama mainly supports English, we use Chatgpt-3.5\footnote{https://chat.openai.com/} to translate non-English problems into English. We only keep the record that the user can pass all test samples for the same problem no more than three times, and choose one at random as the user's answer to the problem. We combined interaction records from problem sets with average user interaction records both below and above 100 to form the PCISparse and PCIDense datasets, respectively. Then we conduct a random split of 8/1/1 by programming problem, validation, and test to avoid data leakage. To prevent privacy disclosure, we have excluded personal and sensitive data such as user names and email addresses, retaining only the unique student IDs as individual identifiers.
\subsection{Baseline Setup Details}
\label{subsec:A2}
We compare our model with mainstream models as follows:\(\bm{(1)}\)\textbf{CodeLlama}: CodeLlama Instruct 7B version can be seen as a benchmark for code generation without personalization, we can observe the degree to which each method succeeds in personalizing the LLM for the target task; \(\bm{(2)}\)\textbf{DAPT}: We perform domain Adaptive Pre-Training~\cite{gururangan2020don} by finetuning on user-specific data; 
\(\bm{(3)}\)\textbf{L-LDB}: it personalizes to a specific software project for personalized unit test generation of Java methods~\cite{zlotchevski2022exploring}. This is achieved by freezing most parameters in the baseline model and only performing lightweight fine-tuning on the last decoder block;
 \(\bm{(4)}\)\textbf{Adapter}: It is used to fine-tune the LLM, and the parameters of the original network remain unchanged, achieving a high degree of parameter sharing~\cite{houlsby2019parameter}.
We split the data in the dataset according to the user, and train a corresponding single-user model with each user's data in turn. The result is calculated by averaging the sum of all users' metrics. For parameter Settings of  Adapter, refer to llama-recipes\footnote{https://github.com/facebookresearch/llama-recipes}. The llama-recipes repository provides fine-tuning code for CodeLlama. In the training stage of baselines, the prompt template for baselines is shown in Fig.~\ref{AppendixFigure-baselines}.
\begin{figure}[!htb]
\centering
\resizebox{\linewidth}{!}{
\fcolorbox{black}{gray!6}{
\parbox{\linewidth}{
\noindent$\bullet$ \textbf{Prompt template.} Give you a programming problem <q>, please provide answers in Java.\\
\noindent$\bullet$ \textbf{Target Response.} <c>
}
}
}
\caption{\label{AppendixFigure-baselines}
The prompt template for baselines.
}
\end{figure}
 
\subsection{Training and Decoding Details}
\label{subsec:A3}
we set \(\tau\) as 0.5 and \(\alpha\) as 0.55. In the first training stage, we set the batch size to 8. In the second training stage, we set the batch size to 4. In both training stages, we used four A800 graphics cards to train the model and truncate the total length of the input statement and output target to a maximum of 2048 tokens, the learning rate is set to 1e-4. 

In the first training of explicit coding style attributes, we choose 25 style attributes, of which 24 attributes are shown in Fig.~\ref{Appendix-guideA}. Some style attributes for CheckStyle checking are shared by all users. These feature differences of users cannot be reflected in the evaluation of personalized code generation. However, during the process of acquiring coding style attributes, leveraging features shared among all users' code assists the LLM in gaining a deeper understanding of these attributes. Therefore, such features are preserved in the residual learning dataset to facilitate the acquisition of coding style attributes. Consequently, we retain ``Indentation" as a fundamental style convention, which means: ``Control indentation between comments and surrounding code.''. Since some semantic coding style features can be explicitly defined by CheckStyle, we also put them into the explicitly coding attributes. Therefore, explicit coding style attributes focus more on learning syntax style, while implicit style features are more concerned with learning semantic style.

For coding style attributes training, we construct a record with the number of style attributes less than or equal to 5 and balance the attribute of each style in the training. The number of records for each style as a residual attribute does not exceed 600 pieces. No more than 75 pieces of data were evaluated for each style attribute. In the experiment for studying the number of attributes, setting the number of attributes to 1 means without using residual learning, we use prompt as shown in Fig.~\ref{appendixFigure-NoResidual}. For implicit style features training, the number of vector representations of implicit style features for each user is set to 5.

\begin{figure}[!htb]
\centering
\resizebox{0.98\linewidth}{!}{%
\fcolorbox{black}{gray!6}{
\parbox{\linewidth}{
\noindent$\bullet$ \textbf{Attribute Prompt without Residual.}
You are given one piece of code <$c$> along with their corresponding style convention <$A_u$>. Please identify and explain the style convention. 

\noindent$\bullet$ \textbf{Target Response.} <text expression of $a_k$> is present in code; the style convention of <text expression of $a_k$> indicates <interpretation of $a_k$>.
}
}
}
\caption{\label{appendixFigure-NoResidual}
The prompt template without residual learning.
}
\end{figure}

\begin{figure}[!htb]
\centering
\resizebox{0.98\linewidth}{!}{%
\fcolorbox{black}{gray!6}{
\parbox{\linewidth}{
\noindent$\bullet$ \textbf{Prompt template 1.}
Here is an incomplete code <$q$>, you need to complete. Wrap your code answer using \textasciigrave\textasciigrave\textasciigrave, your code must include a complete implementation of the 'Solution' class with exactly one function in the class.
\\
\noindent$\bullet$ \textbf{Prompt template 2.}
Give you a piece of Java code, please continue to write the unfinished function <$q$>.
\\
\noindent$\bullet$ \textbf{Prompt template 3.}
Give you a programming question <$q$>, please provide answers in Java.
}
}
}
\caption{\label{appendixFigure-cor}
The prompt templates for the correctness test.
}
\end{figure}
\begin{table*}[!hbt]
\centering
\resizebox{0.95\linewidth}{!}{%
\begin{tabular} {|llcp{0.5\linewidth}|}
\hline 
aspect & criteria & Coding Style & Description\\
\hline
\multirow{4}{*}{ Structure } &NoLineWrap &Syntax&\text{Do not put ‘\}’ on its own line.} \\
&AvoidStarImport  &Syntax&\text{Do not break after ',' but before '.'.} \\
&OneTopLevelClass  &Syntax&\text{break import and package lines.}\\
&EmptyLineSeparator  &Semantic&\text{import statements that use the * notation.} \\
\hline
\multirow{15}{*}{ Formatting } &RightCurly  &Syntax&\text{Do not put ‘\}’ on its own line.} \\
&SeparatorWrap  &Syntax&\text{Do not break after ',' but before '.'.} \\
&WhitespaceAround  &Syntax&Do not use a space between a reserved word and its follow-up bracket,\textit{e.g.}, if(. \\
&GenericWhitespace  &Syntax&Use a space before the definition of generic type, \textit{e.g.}, List <.\\
&OperatorWrap  &Syntax&\text{Break after '=' but after other binary operators.}\\
&LineLength  &Semantic&\text{The line length exceeds 100 characters.}\\
&LeftCurly  &Syntax&\text{Do not put '\{' on the same line of code.}\\
&EmptyBlock  &Syntax&\text{Have empty block for control statements.}\\
&NeedBraces  &Syntax&\text{Do not use braces for single control statements.}\\
&MultipleVariableDeclarations  &Syntax&Not every variable declaration is in its own statement and on its own line.\\
&OneStatementPerLine  &Syntax&\text{there is not only one statement per line.}\\
&UpperEll  &Syntax&long constants are defined with an upper ell. That is 'l' and not 'L'.\\
&ModifierOrder  &Syntax&Do not follow the order: public, protected, private, abstract, default, static, final, transient, volatile, synchronized, native, strictfp.\\
&FallThrough  &Semantic&Do not put a fall-through comment in a switch If a 'case' has no break, return, throw, or continue.\\
&MissingSwitchDefault  &Semantic&switch statement does not has a default clause.\\
\hline
\multirow{5}{*}{ Naming }  &TypeName &Syntax&Type name is not in UpperCamelCase.\\
&MethodName  &Syntax&Method name is not in lowerCamelCase. \\
&MemberName  &Syntax& Member name is not in lowerCamelCase.\\
&ParameterName  &Syntax&Parameter name is not in lowerCamelCase.\\
&LocalVariableName  &Syntax&Local variable name is not in lowerCamelCase. \\
\hline
\end{tabular}
}
\caption{\label{Appendix-guideA}
The 24 criteria for characterizing the coding style.
}
\end{table*}

\subsection{Details of Human Study}
\label{subsec:A4}
We conduct a user study to verify the effectiveness of our CSS metric for evaluating coding styles compared with humans. In particular, we compare MPCoder with L-LDB and Adapter by conducting a user study on PCIDense dataset. We asked 5 users to answer questionnaires of 60 comparative questions, totaling 300 answers. All of the users are majored in Computer Science and/or Software Engineering and with more than 4 years of Java programming experience.

\begin{figure*}[!hbt]
\centering
\includegraphics[width =\textwidth]{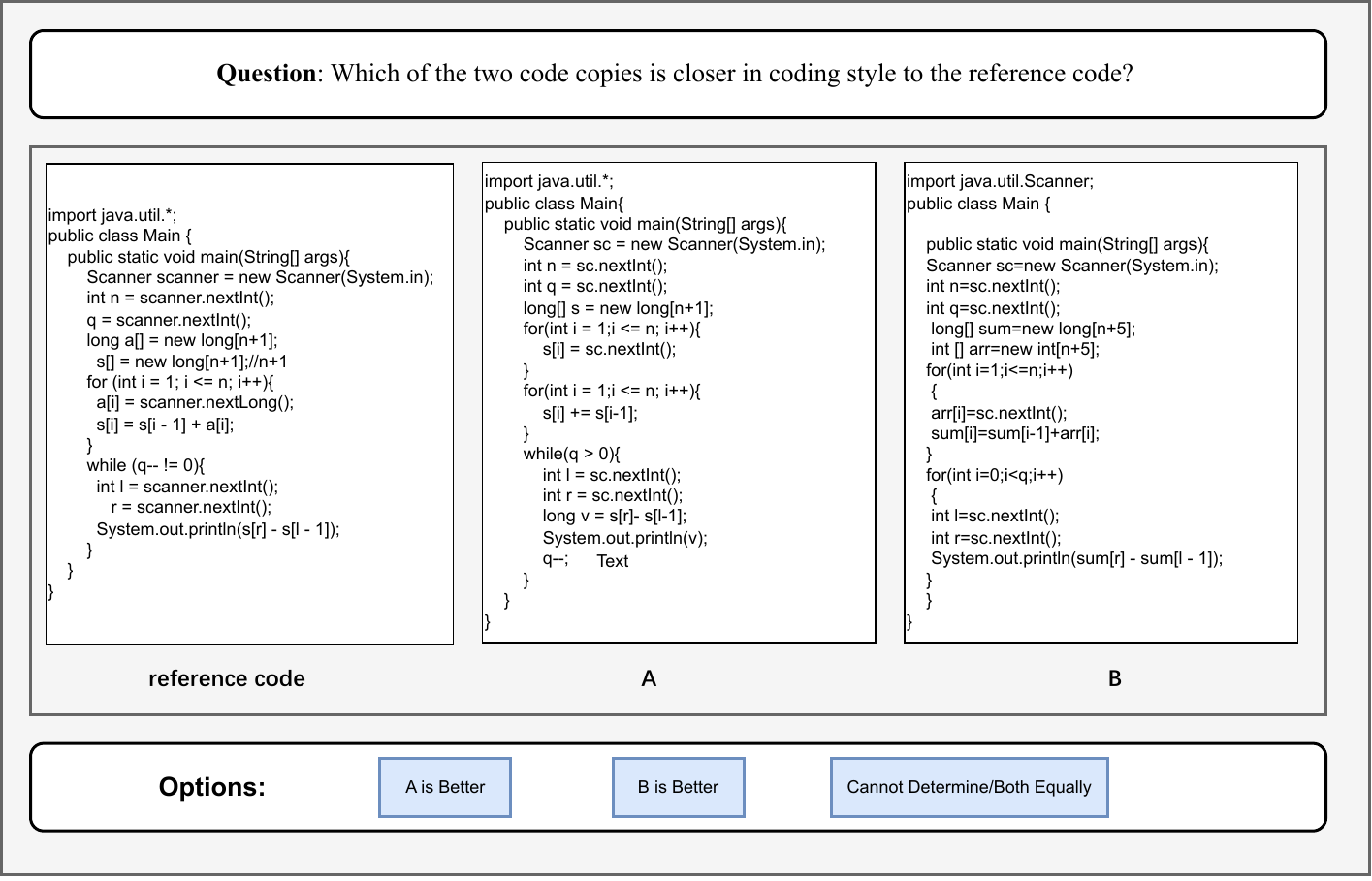}
\caption{ An example of the human study.
} 
\label{fig:UserStudyExample}
\end{figure*}

Prior to the study, users are informed in the definition of syntax and semantic coding style. Syntactic style includes common formats and structures; semantic style includes aspects such as data flow and meaningful naming. Samples are randomly selected from a pool of test data. To simplify the decision-making process for users, all questions present them with a choice between two options. Users are asked to answer the question: “Which of the two code copies is closer in coding style to the reference code?”, and every user is provided with three options (i.e., A is Better, B is Better, Cannot Determine/Both Equally). It is worth mentioning that the users do not know which code snippet is generated by which method. we provide an example of the question as shown in Fig~\ref{fig:UserStudyExample}.

\subsection{Details for Adapting to New Users}
\label{subsec:A5}
\paragraph{Scenario 1.}   In this scenario, we can utilize both full training and incremental training to update the model. For the incremental training approach, we freeze the model parameters and only train the implicit style features of the new user based on his historical records. Due to time constraints, we have conducted a preliminary verification comparing the results of full training versus incremental training for a single new user in Table~\ref{ScenarioTable}. Although the effectiveness of incremental training is slightly inferior to that of full training, this approach avoids the need to retrain the entire model, thus showcasing the feasibility and efficiency of our proposed approach for adding new users.
\begin{table}[!thb]
\resizebox{0.48\textwidth}{!}{
\begin{tabular}{lcccc}
\toprule 
\multirow{1}{*}{\textbf{Method}} & \multirow{1}{*}{\textbf{CSS}} & \multirow{1}{*}{\textbf{BLUE} } & \multicolumn{1}{c}{\textbf{Rouge-1}} & \multirow{1}{*}{\textbf{Rouge-2} } \\
\midrule 
Full training & 60.92 & 63.39 & 52.21 & 39.25\\
Incremental training & 58.21 & 63.35 & 51.65 & 38.28\\
\bottomrule
\end{tabular}
}
\vspace{-0.5em}
\caption{Incremental training on PCIDense. }
\label{ScenarioTable}
\vspace{-1em}
\end{table}

\paragraph{Scenario 2.} In this scenario, we can use {\sc MPCoder}\textsubscript{ESF} which only utilizes explicit style attributes for inference. Since explicit attributes do not depend on specific user data, the user only needs to specify the corresponding explicit coding style attribute or opt for a default setting. Consequently, the model can generate code that aligns with the user's syntax style. The experimental results are shown in the table~\ref{citation2-guide}. Although {\sc MPCoder}\textsubscript{ESF} may not be as efficient as {\sc MPCoder} and is focused primarily on syntax style, it does not require any training for new users.
\begin{figure*}[!thb]
\centering
\includegraphics[width =\textwidth]{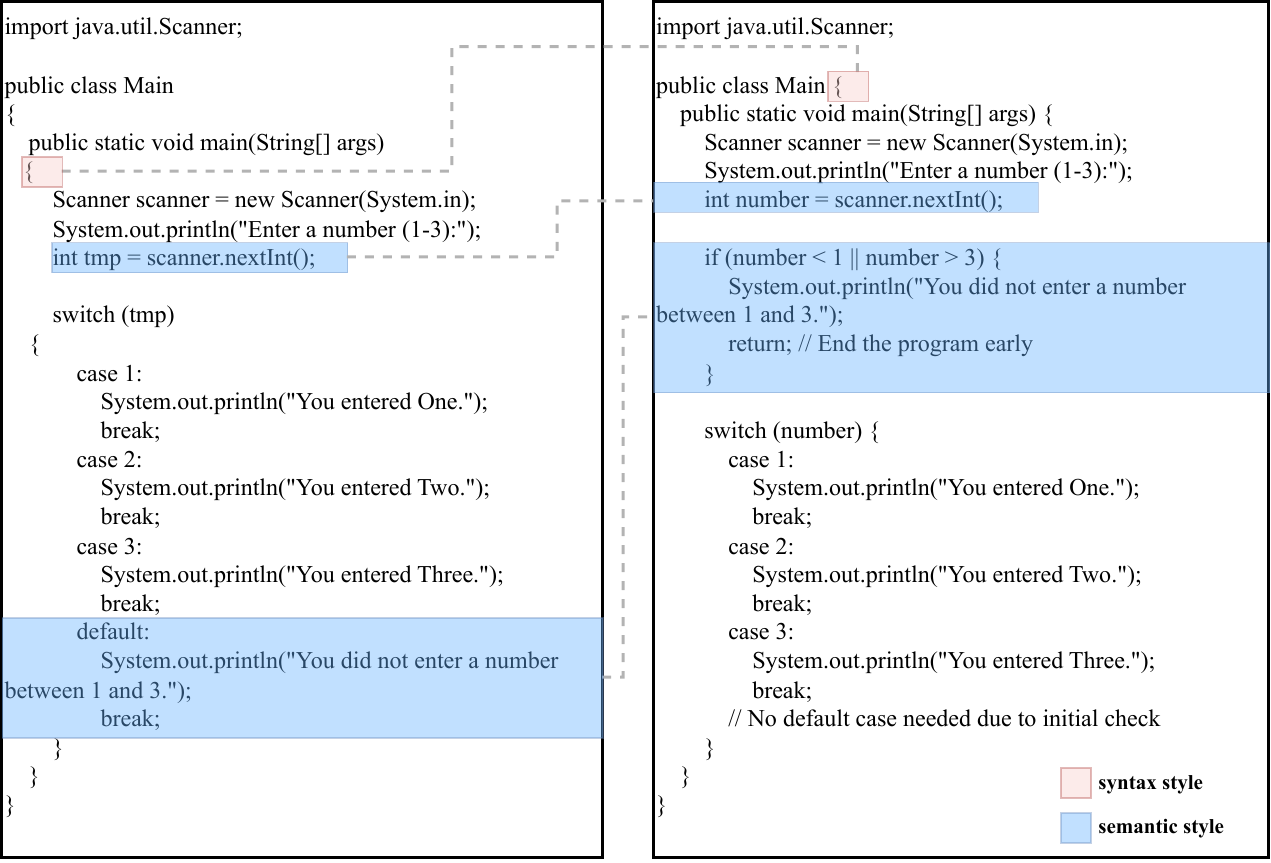}
\caption{Syntax and Semantic Coding Styles.
} 
\label{fig:IEexample}
\end{figure*}

\section{Evaluation Statement}
\subsection{Correctness Evaluation}
\label{subsec:B1}
Models trained on PCISparse and PCIDense are evaluated on a dataset of human-x code correctness tests. The correctness of the first reply code of the model is tested by greedy decoding. On PCIDense dataset, we fully test all problems in HumanEval-X for each user and report the average values based on three prompt templates. Because there are too many users in PCISparse data, we randomly select 50 users as the object of correctness verification. The prompt template for the correctness test is shown in Fig.~\ref{appendixFigure-cor}.
\subsection{Java Style Criteria}
\label{subsec:B2}
As shown in Table~\ref{Appendix-guideA}, we select 24 code criteria that can reflect the coding style of Java programming problems from three aspects: structure, naming, and formatting. It is important to note that the proposed CSS metric incorporates 20 style criteria for identifying syntax style and 4 style criteria for identifying semantic style. For example, the attributes ``FallThrough'' and ``MissingSwitchDefault'' are utilized to detect differences in the code execution order. Both of which pertain to the semantic style of control flow and data flow~\cite{li2022ropgen}. Specifically, ``FallThrough'' means ``Do not put a fall-through comment in a switch If a 'case' has no break, return, throw, or continue''; ``MissingSwitchDefault'' means ``switch statement does not have a default clause''. In other words, the code style attributes checked by CheckStyle contain both syntax features and semantic features, and our CSS evaluation metric using these features can estimate model performance from both syntactic and semantic perspectives.

\begin{figure*}[!thb]
\centering
\includegraphics[width =\textwidth]{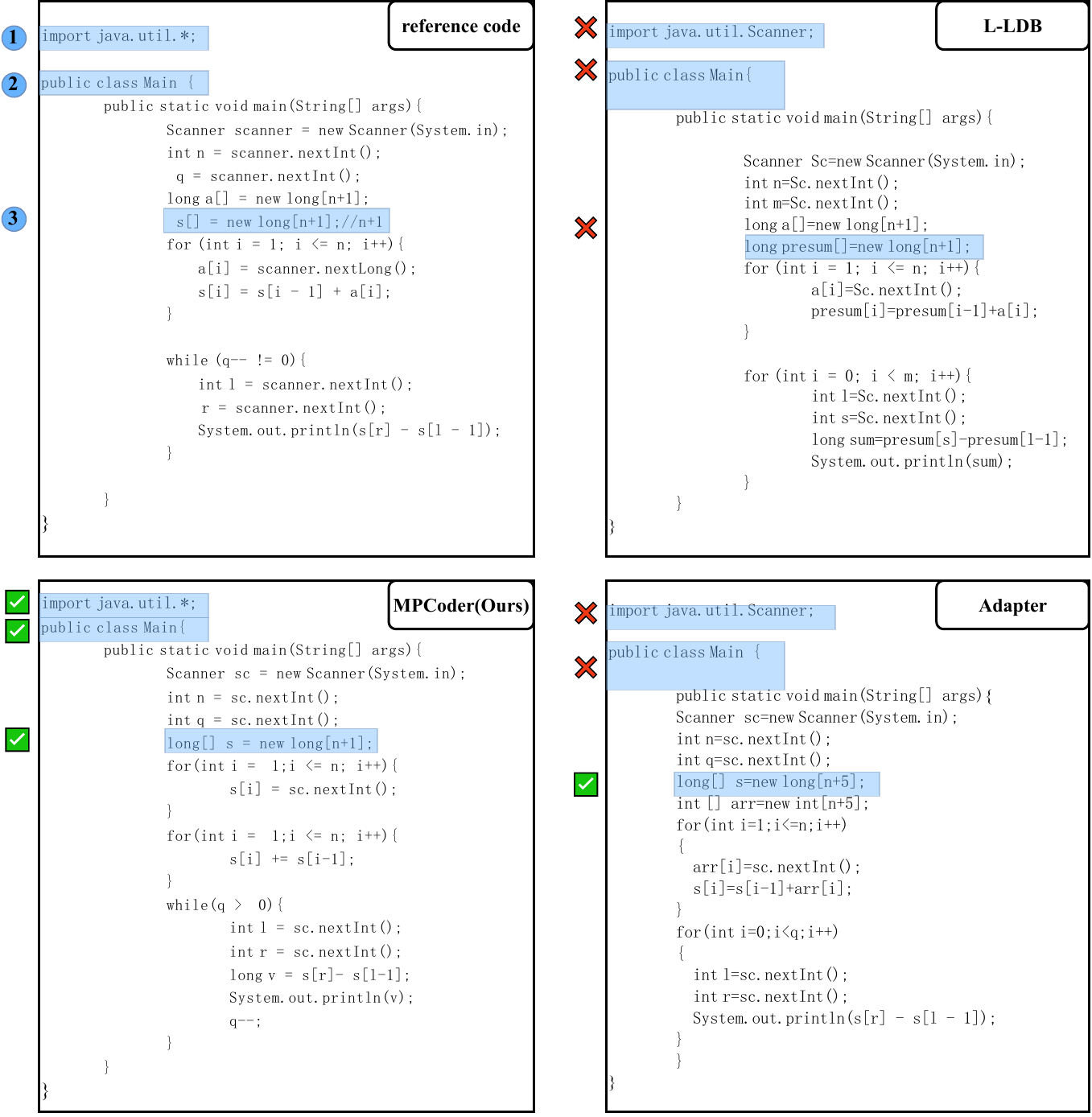}
\caption{Syntax and Semantic Coding Styles.
} 
\label{fig:modelperformance}
\end{figure*}

\subsection{Syntax and Semantic Coding Styles}
\label{subsec:B3}

Fig.~\ref{fig:IEexample} shows an example of the syntax and semantic differences in coding styles. Both copies of the code solve the same problem, but the code reflects different syntax and semantic styles.
\paragraph{Syntax style differences.} 
The curly braket "\{" in the left code copy is placed on the same line as the preceding statement, while the right code copy places it on a separate new line, which reflects the layout style difference in the format.
\paragraph{Semantic style differences.} 
The left code copy uses "tmp" as a temporary variable name, while the right code copy uses "number" as a numerical variable name, reflecting different semantic styles in naming conventions. The left code copy uses default in the Switch statement, while the right code copy does not use default. Due to the different code control flows, the order of actual program execution may also be different. The left code may run to the end of the default statement, and the right code could output and terminate the program at the beginning. The two pieces of code represent different data flows, or design patterns, that reflect the semantic coding styles of different developers.

\subsection{Example of Model Performance}
\label{subsec:B4}
Figure~\ref{fig:modelperformance} shows the personalized code generation results of the same question for the same user regarding different models. Significant coding styles are highlighted in blue, showing that the generation result of our model aligns more closely with the overall style (including syntax style and semantic style) of the reference code than those of Adapter and L-LDB.

\end{document}